\pdfoutput=1
\documentclass{article}
\usepackage[preprint]{corl_2022}

\usepackage{enumitem}
\usepackage{footmisc}
\usepackage{graphicx}
\graphicspath{{figs/}}
\DeclareGraphicsExtensions{.pdf,.jpeg,.png}

\usepackage{amsmath}
\usepackage[usenames,dvipsnames]{xcolor}

\usepackage[caption=false,font=normalsize,labelfont=sf,textfont=sf]{subfig}
\usepackage{tikz}
\usepackage{tikzscale}
\usetikzlibrary{arrows,backgrounds,decorations,decorations.pathmorphing,positioning,fit,automata,shapes,patterns,plotmarks,calc}
\usetikzlibrary{positioning}
\usepackage{stfloats}
\usepackage{url}
\usepackage[ruled, vlined, linesnumbered]{algorithm2e}

\usepackage{macros}
\usepackage{booktabs}
\tikzset{mypic/.style={scale=1.0, transform shape, node distance={9mm}},
	roundnode/.style={circle, draw=teal!60, fill=teal!5, very thick, minimum size=5mm},
	squarednode/.style={rectangle, draw=red!60, fill=red!5, very thick, minimum size=5mm},
}

\newcommand{\hamm}{\textit{Hamming} }

\newcommand{\appref}[1]{Appendix \ref{#1}}

\title{Learning Performance Graphs from Demonstrations \\ via Task-Based Evaluations}

\author{
  Aniruddh G.~Puranic, Jyotirmoy V.~Deshmukh and Stefanos Nikolaidis\\
  Department of Computer Science\\
  University of Southern California, USA\\
  \texttt{\{puranic, jdeshmuk, nikolaid\}@usc.edu} \\
}

\begin{document}
% make the title area
\maketitle

\begin{abstract}
In the paradigm of robot learning-from-demonstrations (LfD), understanding and evaluating the demonstrated behaviors plays a critical role in extracting control policies for robots. Without this knowledge, a robot may infer incorrect reward functions that lead to undesirable or unsafe control policies. Prior work has used temporal logic specifications, manually ranked by human experts based on their importance, to learn reward functions from imperfect/suboptimal demonstrations. To overcome reliance on expert rankings, we propose a novel algorithm that learns from demonstrations, a partial ordering of provided specifications in the form of a performance graph. Through various experiments, including simulation of industrial mobile robots, we show that extracting reward functions with the learned graph results in robot policies similar to those generated with the manually specified orderings. We also show in a user study that the learned orderings match the orderings or rankings by participants for demonstrations in a simulated driving domain. These results show that we can accurately evaluate demonstrations with respect to provided task specifications from a small set of imperfect data with minimal expert input. 
\end{abstract}


\section{Introduction}
\label{sec:intro}
In human-robot interaction, understanding the behaviors exhibited by humans and robots plays a key role in robot learning, improving task efficiency, collaboration and mutual trust~\citep{naesm_humanAI,gazit_fmhri2021,gombolay_xai,sanneman2022}. Demonstrated behaviors can be evaluated by specifying a cumulative reward function that assigns behaviors a numeric value; the implicit assumption here is that higher cumulative rewards indicate good behaviors. Such cumulative rewards are then used with reinforcement learning (RL) to learn an optimal policy. Poorly-designed reward functions can pose the serious risk of reward hacking where the agent behavior maximizing the cumulative reward is undesirable or unsafe~\citep{amodei2016concrete}. An alternative method is the use of high-level task descriptions using logical specifications (rather than manually defined rewards), and approaches that use task descriptions expressed in temporal logic are quite popular in robotics~\citep{gundana_stl,li_reinforcement_2017,belta_lfd,aksaray_q-learning_2016,wen_2015}. Of particular relevance to this paper is the use of the specification language Signal Temporal Logic (STL) \citep{stl_complexity}. Recent work develops the LfD-STL framework \citep{puranic_corl2020,puranic_ral2021}; this framework uses STL specifications to infer rewards from user-demonstrations (some of which could be suboptimal or unsafe) and has shown to outperform inverse reinforcement learning (IRL) methods \citep{mce_irl_ziebart, suay_aamas16, adv_irl} in terms of number of demonstrations required and the quality of rewards learned.

In the LfD-STL framework, STL specifications describe desired objectives, and demonstrations show how to achieve these objectives. For example, in an autonomous driving scenario, an STL specification could describe the task ``reach the goal while avoiding obstacles''. STL is equipped with quantitative semantics that indicate how well the demonstrations performed on the tasks, which are used to infer rewards. The inferred rewards guide the robotic agent towards policies that produce desirable behaviors. Typically, robotic systems are difficult to characterize using a single specification, and users may thus seek for policies that satisfy several task specifications. Not all specifications may be equally important; for example, a hard safety constraint is more important than a performance objective. LfD-STL permits users to thus provide several specifications, but also requires them to manually specify their preferences or priorities on specifications. These preferences are then encoded in a directed acyclic graph (DAG) that, in this paper, we call the {\em performance graph}. Based on a given performance graph and the quantitative semantics of each STL specification, the LfD-STL framework then defines a state-based reward function that is used with off-the-shelf RL methods to extract a policy. This framework offers few crucial advantages: (i) STL allows defining non-Markovian rewards useful for sequential, patrolling, and reactive tasks, (ii) applications to high-dimensional, continuous and stochastic spaces, and (iii) has empirically demonstrated significant reductions in sample complexity for learning.

\begin{figure}[tb]
\includegraphics[width=\linewidth]{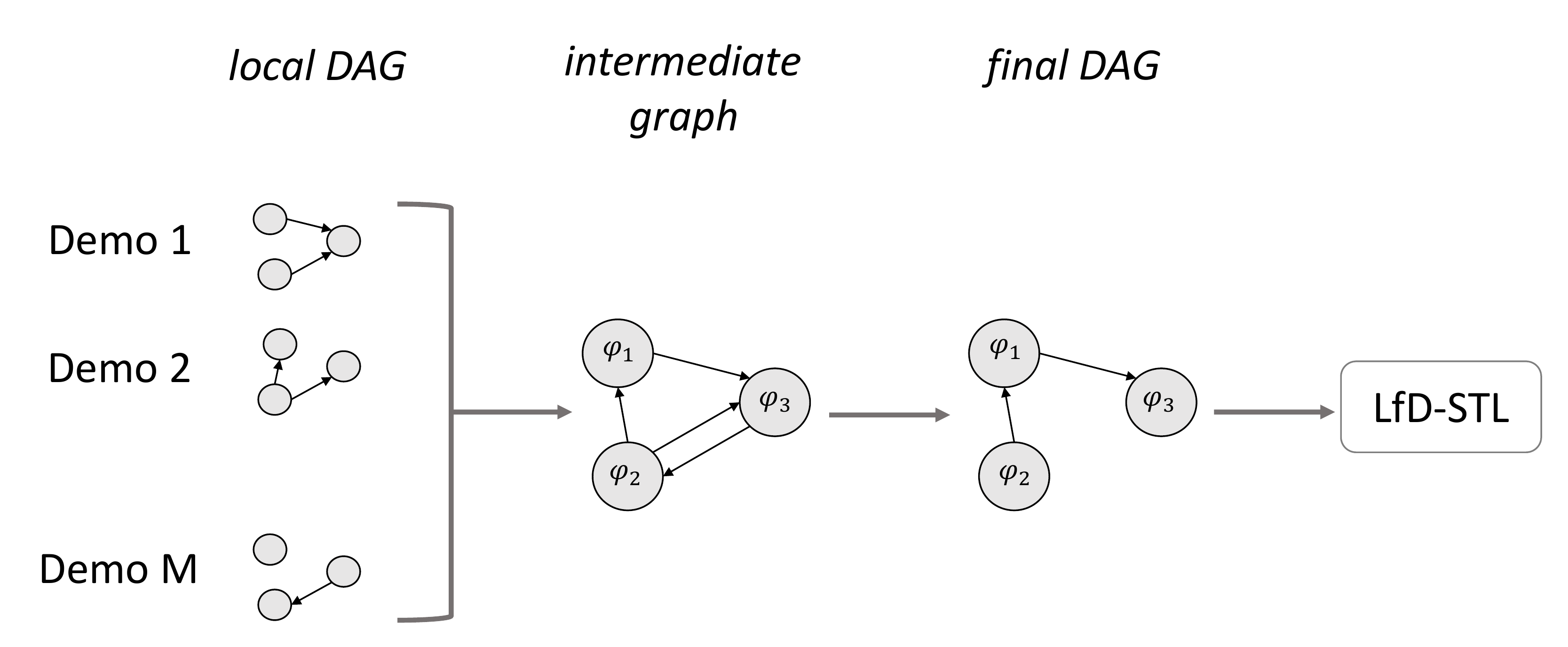}
\caption{Overview of PeGLearn algorithm.}
\label{fig:dag_framework}
\end{figure}

A key limitation of the LfD-STL framework is that the onus on providing the performance graph is on the user, which becomes infeasible when there are numerous task specifications. Moreover, there could exist multiple ways of performing a task and a challenge in LfD is whom the agent should imitate, i.e., disambiguate the demonstrations. We propose solutions to these problems by using the STL specifications and quantitative semantics to not only evaluate the performance of demonstrations, but also to use the evaluations to infer the performance graph. We propose the Performance-Graph Learning (PeGLearn) algorithm that systematically constructs DAGs from demonstrations and STL evaluations, resulting in a representation that can be used to explain the performance of provided demonstrations. A high-level overview of this algorithm used in conjunction with prior LfD framework is shown in \autoref{fig:dag_framework}, which we discuss in detail in \autoref{sec:framework}.

In complex environments where it is non-trivial to express tasks in STL, we use human annotations (ratings or scores) of the data. Examples of complex tasks in human-robot interactions can include descriptions like ``tying a knot'' or having ``fluency in motion'' in robotic surgery, where even experts can struggle to express the task in formal logic, or specifying the task formally may not be expressible in a convenient logical formalism. In our setting, rating scales can replace temporal logics by (i) choosing queries that assess performance and (ii) treating the ratings/scores as quantitative assessments\footnote{Here we assume that $Likert$ scales are interval scales\citep{norman2010likert}.}. There is precedence of such quantitative assessments; for example, $Likert$ ratings from humans are used as ground-truth measurements of trust~\citep{chen_niko_trust2018}. We perform several simulation experiments in autonomous driving and complex navigation of an industrial mobile robot (MiR100) to validate our approach. We also discuss the complexity of PeGLearn and draw comparisons with existing learning techniques.

\section{Background}
\label{sec:prelims}

In this section, we provide various definitions and notations used in our methodology and experiments.

\begin{definition}[Demonstration]
	A demonstration is a finite sequence of state-action pairs in an environment that is composed of a state-space $S$ and action-space $A$ that can be performed by the agent. Formally, a demonstration $\demo$ of finite length $L \in \mathbb{N}$ is $\demo =\langle(s_1, a_1), (s_2, a_2), ..., (s_L, a_L)\rangle$, where $s_i \in S$ and $a_i \in A$. That is, $\demo$ is an element of $(S \times A)^L$. We interchangeably refer to demonstrations as trajectories. \label{def:demo}
\end{definition}

Each environment is governed by tasks or objectives, which we refer to as specifications or requirements, denoted by $\varphi$. In this regard, we define a rating function via \autoref{def:ratingfn}. 

\begin{definition}[Rating Function]
	A \textit{rating function} $\mathcal{R}$ is a real-valued function that maps a specification and a time-series data or trajectory to a real number, i.e., $\mathcal{R}: \Phi \times \demoset \rightarrow \Reals$, where, $\Phi$ and $\demoset$ are each infinite sets of specifications and demonstrations, respectively.
\label{def:ratingfn}
\end{definition}

Intuitively, the rating function describes how ``well'' the specifications are met (satisfied) by a trajectory. The rating function can be obtained via the quantitative (robustness) semantics in temporal logics \citep{peglearn_supp} or human ratings via surveys, annotations, etc. It indicates the score or signed distance of the time-series data to the set of temporal data satisfying a specification. For a given specification $\varphi$ and a demonstration $\demo$, the \textit{rating} (also referred to as \textit{evaluation} or \textit{score}) of $\demo$ with respect to $\varphi$ is denoted by $\rho = \mathcal{R}(\varphi, \demo)$. This $\rho$ is negative if $\demo$ violates $\varphi$, and non-negative otherwise.

\begin{definition}[Directed Acyclic Graph]
	A directed acyclic graph or DAG is an ordered pair $G = (V, E)$ where $V$ is a set of elements called vertices or nodes and $E$ is a set of ordered pairs of vertices called edges or arcs. An edge $e = (u, v)$ is directed from a vertex $u$ to another vertex $v$. \label{def:dag}
\end{definition}

A path $x \leadsto y$ in $G$ is a set of vertices starting from vertex $x$ and ending at vertex $y$ by following the directed edges from $x$. Each vertex $v \in V$ is associated with a real number - weight of the vertex, represented by $w(v)$. Similarly, each edge $(u, v) \in E$ is associated with a real number - weight of the edge and is represented by $w(u, v)$. Notice the difference in the number of arguments in the notations of vertex and edge weights. Similar to prior work \citep{puranic_corl2020, puranic_ral2021}, we define each node $v \in V$ of some DAG, $G$, to represent a task specification.

\section{Problem Definition and Methodology}
\label{sec:method}

\subsection{Problem Formulation}
\label{sec:problem_definition}
To accomplish a set of tasks, we are given: (i) a finite dataset of $m$ demonstrations $\demoset = \{\demo_1, \demo_2, \cdots, \demo_m\}$ in an environment, where each demonstration is defined as in \autoref{def:demo}, (ii) a finite set of $n$ specifications $\Phi = \{\varphi_1, \varphi_2, \cdots, \varphi_n\}$ to express the high-level tasks and by which a set of scores for each demonstration evaluated on each of the $n$ specifications $\boldsymbol{\rho_\demo} = [\rho_1, \cdots, \rho_{\abs{\Phi}}]^T$ is obtained. We can then represent this as an $m \times n$ matrix $\mathcal{Z}$ where each row $i$ represents a demonstration and a column $j$ represents a specification. An element $\rho_{ij}$ indicates the rating or score of demonstration $i$ for specification $j$, i.e., $\rho_{ij} = \mathcal{R}(\varphi_j, \demo_i)$. 

\begin{equation}
        \mathcal{Z} = \begin{bmatrix}
        \rho_{11} & \rho_{12} & \cdots & \rho_{1n} \\
        \rho_{21} & \rho_{22} & \cdots & \rho_{2n} \\
        \vdots & \cdots & \ddots & \vdots \\
        \rho_{m1} & \rho_{m2} & \cdots & \rho_{mn} \\
        \end{bmatrix}
        = 
        \begin{bmatrix}
        \boldsymbol{\rho_{\demo_1}^T} \\
        \boldsymbol{\rho_{\demo_2}^T} \\
        \vdots \\
        \boldsymbol{\rho_{\demo_m}^T} \\
        \end{bmatrix}
    \label{eqn:total_demo_rob}
\end{equation}

As in prior work on LfD-STL, we need to compute a cumulative score or rating $r_\demo$ for each demonstration to collectively represent its individual specification scores, and so we have an $m \times 1$ vector $\mathbf{r} = [r_{\demo_1}, r_{\demo_2}, \cdots, r_{\demo_m}]^T$. To obtain the cumulative scores, we require a scalar quantity or \textit{weight} associated with each specification indicating its (relative) priority/preference/importance over other specifications. We thus have a weight vector $\mathbf{w} = [w_1, w_2, ..., w_{\abs{\Phi}}]^T$ from which we can obtain the cumulative scores as $\mathcal{Z} \cdot \mathbf{w} = \mathbf{r}$. In other words, for each demonstration $\demo$, $r_\demo = \boldsymbol{\rho_\demo}^T \cdot \mathbf{w}$. The objective is to compute both $\mathbf{w}$ and $\mathbf{r}$, given only $\mathcal{Z}$, such that the ``better'' demonstrations have higher cumulative scores than others and ranked appropriately by the LfD-STL framework, i.e., generate {\em partial order}; this is an unsupervised learning problem. One of the approaches to computing $\mathbf{w}$ proposed in \citep{puranic_corl2020, puranic_ral2021} required the demonstrators to specify their preferences encoded as a DAG and the weights were computed via \eqref{eqn:dag_weights}\footnote{Before robustness values of various temporal logic specifications are combined, they are normalized using $\mathtt{tanh}$ or piece-wise linear functions.\label{fn:norm_rob}}.
\begin{equation}
    \label{eqn:dag_weights}
    w(\varphi) = |\Phi| - |ancestor(\varphi)|
\end{equation}
where, $ancestor(v) = \{u \mid u \leadsto v, u \in V\}$, i.e., the ancestors of a vertex $v$ is the set of all vertices in $G$ that have a path to $v$. However, this is not data-driven as it requires human inputs to define the weights and is only feasible if the number of specifications is small. To overcome this, we can rely on data-driven approaches such as unsupervised learning. In our experiments, we show how existing methods are inefficient, and we thus propose a new approach by learning a DAG directly from demonstrations (i.e., without human inputs) and using \eqref{eqn:dag_weights} to compute weights for the LfD-STL framework to extract rewards for RL. The DAG contains the elements of $\Phi$ as its vertices, and the relative differences in performance between specifications as edges. We refer to this as the Performance-Graph since it captures the performance of the demonstrations w.r.t. the task specifications. This final graph is required to be acyclic so that topological sorting can be performed on the graph to obtain an ordering of the nodes and hence specifications, i.e., topological ordering does not apply when there are cycles in the graph.

An assumption that we make is that, at least 1 demonstration satisfies all the specifications of $\Phi$, but does not have to be optimal (i.e., having the highest rating) w.r.t. those specifications; we argue that this is a reasonable assumption to make compared to related LfD works like IRL that require a large sample size of nearly-optimal (i.e., close to the highest rating) demonstrations and also to show that the task(s) can be realized, even if suboptimally, under the given specifications. The last assumption is that the demonstrators' intentions are accurately reflected (in terms of performance) in the demonstrations provided, therefore we consider each demonstration equally important when inferring graphs.  

\subsection{Rating-based Graph}
\label{sec:framework}
In this section, we describe the procedure to create the Performance-Graph from ratings or scores obtained either automatically by temporal logics or provided by human annotators. This process involves 3 main steps: 
\begin{enumerate}[leftmargin=1em,nosep]

\item Constructing a {\em local} weighted-DAG for each demonstration based on its individual specification scores.

\item Combining the $local$ graphs into a single weighted directed graph, which is not necessarily acyclic as it can contain bidirectional edges between nodes.

\item Converting the resultant graph into a weighted DAG.

\end{enumerate}

The framework in \autoref{fig:dag_framework} depicts the 3 steps described above and the final stage where the inferred DAG is fed to the LfD-STL framework to learn rewards and perform RL.

\subsection{Generating \textit{local} graphs}
\label{sec:local_graph}
Each demonstration $\demo \in \demoset$ is associated with a vector of ratings $\boldsymbol{\rho_\demo} = [\rho_1, ..., \rho_{\abs{\Phi}}]^T$, and the objective is to construct a weighted DAG for $\demo$ from these evaluations. We propose the novel \autoref{alg:peglearn_local}, where, initially, the evaluations are sorted in non-increasing order with ties broken arbitrarily (lines 3--5). This creates a partial ordering based on the performance of the demonstrations regarding each specification, and is represented by a DAG to capture the partial ordering. Though DAGs can be represented by either adjacency lists or adjacency matrices, in this work, we represent them using adjacency matrices for notational convenience. 

\begin{algorithm}[htbp]
% \SetAlgoLined
\DontPrintSemicolon
\SetKwInput{Input}{Input}
\Input{$\demo$ := a demonstration of any length $L$; $\Phi$ := set of $n$ specifications; $\epsilon$: threshold (tunable)
} 
\KwResult{Constructs local Performance-Graph $G_\demo$}
\BlankLine

\Begin{
	$G_\demo \leftarrow \mathbf{0_{n \times n}}$ \tcp*[r]{zero matrix}
	$\mathcal{S} \leftarrow \emptyset$ \tcp*[r]{Create an empty queue}
	\For{$j=1$ to $n$}{
		Obtain the rating or score $s_j$ for specification $j$;
		$\mathcal{S}.insert(\langle j,  s_j \rangle)$ \;
	}
	\tcp{Resulting $S$ is an $n \times 2$ matrix where each row is $\langle index, score \rangle$}
	$S' \leftarrow$ sort $\mathcal{S}$ in non-increasing order of scores \tcp*[r]{original indices are recorded}
	\For(\tcp*[h]{no self-loops}){$k=1$ to $n-1$}{
		$\varphi \leftarrow S'[k, 1]$ \tcp*[r]{get index}
		$v \leftarrow S'[k, 2]$ \tcp*[r]{get score}
		\For{$j=k+1$ to $n$}{
			$\varphi' \leftarrow S'[j, 1]$ \;
			$v' \leftarrow S'[j, 2]$ \;
			\If{$(v - v') \ge \epsilon$}{
				$G_\demo[\varphi, \varphi'] \leftarrow G_\demo[\varphi, \varphi'] + v - v'$\;
			}
		}
	}
\Return $G_\demo$ \;
}
\caption{Algorithm to compute local DAG for a single demonstration. \label{alg:peglearn_local}}
\end{algorithm}

\begin{figure}[tb]
\centering
\subfloat[Step 1\label{fig:local_step1}]{
	\begin{tikzpicture}[mypic,
	selectednode/.style={circle, draw=purple!60, fill=purple!5, very thick, minimum size=7mm}]
		%Nodes
		\node[selectednode](phi1){$\varphi_1$};
		\node[roundnode](phi2)[right=of phi1]{$\varphi_2$};
		\node[roundnode](phi3)[below=of phi1]{$\varphi_3$};
		\node[roundnode](phi4)[below=of phi2]{$\varphi_4$};
		
		%Lines
		\draw[->] (phi1.east) -- node[midway, above, sloped] {$\delta_{12}$} (phi2.west);
		\draw[->] (phi1.south) -- node[midway, above, sloped] {$\delta_{13}$} (phi3.north);
%		\draw[->] (phi1) -- node[midway, above, sloped] {$\delta_{14}$} (phi4);
	\end{tikzpicture}
} \hfil
\subfloat[Step 2]{
	\begin{tikzpicture}[mypic,
	selectednode/.style={circle, draw=purple!60, fill=purple!5, very thick, minimum size=7mm}]
		%Nodes
		\node[roundnode](phi1){$\varphi_1$};
		\node[selectednode](phi2)[right=of phi1]{$\varphi_2$};
		\node[roundnode](phi3)[below=of phi1]{$\varphi_3$};
		\node[roundnode](phi4)[below=of phi2]{$\varphi_4$};
		
		%Lines
		% From 1 to 2, 3 and 4
		\draw[->] (phi1.east) -- node[midway, above, sloped] {$\delta_{12}$} (phi2.west);
		\draw[->] (phi1.south) -- node[midway, above, sloped] {$\delta_{13}$} (phi3.north);
%		\draw[->] (phi1) -- node[midway, above, sloped, pos=0.8] {$\delta_{14}$} (phi4);
		
		% From 2 to 3 and 4
		\draw[->] (phi2) -- node[midway, above, sloped, pos=0.2] {$\delta_{23}$} (phi3);
		\draw[->] (phi2.south) -- node[midway, above, sloped] {$\delta_{24}$} (phi4.north);
		
	\end{tikzpicture}
} \hfil
\subfloat[Step 3]{
	\begin{tikzpicture}[mypic,
	selectednode/.style={circle, draw=purple!60, fill=purple!5, very thick, minimum size=7mm}]
		%Nodes
		\node[roundnode](phi1){$\varphi_1$};
		\node[roundnode](phi2)[right=of phi1]{$\varphi_2$};
		\node[selectednode](phi3)[below=of phi1]{$\varphi_3$};
		\node[roundnode](phi4)[below=of phi2]{$\varphi_4$};
		
		%Lines
		% From 1 to 2, 3 and 4
		\draw[->] (phi1.east) -- node[midway, above, sloped] {$\delta_{12}$} (phi2.west);
		\draw[->] (phi1.south) -- node[midway, above, sloped] {$\delta_{13}$} (phi3.north);
%		\draw[->] (phi1) -- node[midway, above, sloped, pos=0.8] {$\delta_{14}$} (phi4);
		
		% From 2 to 3 and 4
		\draw[->] (phi2) -- node[midway, above, sloped, pos=0.2] {$\delta_{23}$} (phi3);
		\draw[->] (phi2.south) -- node[midway, above, sloped] {$\delta_{24}$} (phi4.north);
		
		% From 3 to 4
		\draw[->] (phi3.east) -- node[midway, above, sloped] {$\delta_{34}$} (phi4.west);
	\end{tikzpicture}
} \hfil
\caption{Example $local$ graph for a demonstration.}
\label{fig:local_graph}
\end{figure}
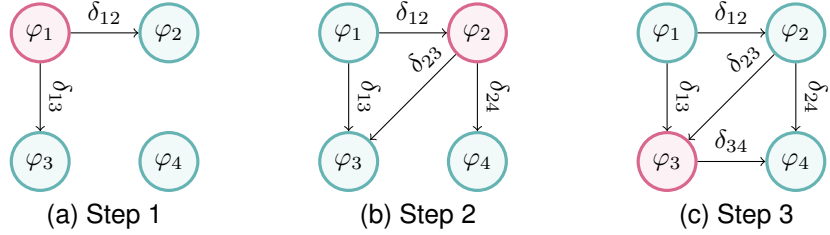

Consider 4 specifications $\varphi_i; i \in \{1, 2, 3, 4\}$. Let a demonstration, say $\demo \in \demoset$ have evaluations $\boldsymbol{\rho_\demo} = [\rho_1, \rho_2, \rho_3, \rho_4]$ and without loss of generality, let them already be sorted in non-increasing values, i.e., $\rho_i \ge \rho_j; \forall i < j$. This sorting is performed in the first \texttt{for} loop of \autoref{alg:peglearn_local}. Recall that each node of the DAG represents a specification of $\Phi$, i.e., a node contains the index of the specification it represents. An edge between two nodes $\varphi_i$ and $\varphi_j$ is created when the difference between their corresponding evaluations is greater than a small threshold value (lines 6--14). This edge represents the relative rating or performance difference between the specifications and creates a partial order indicating this difference. The threshold $\epsilon$ acts as a high-pass filter and can be tuned depending on the normalization of ratings. The intuition is that demonstrations having similar states or features will have similar evaluations for the specifications, and should produce the same partial ordering of specifications. That is, an edge is created if the evaluations differ greatly, e.g., two specifications producing ratings say, 1.0 and 0.99, are numerically different, but have similar performance, so they should be equally ranked (have no edge between them). Without this filter, an edge with a very small weight would be created between them, thereby inadvertently distinguishing similar performances. Formally, $e(\phix{i},\phix{j})$ is added when $\delta_{ij} = (\rho(\phix{i}) - \rho(\phix{j})) \ge \epsilon$\footnote{The demonstration $\demo$ argument in $\rho$ was dropped for convenience since we are only considering 1 demonstration at a time.}. We repeat this process for each node in the DAG (\autoref{fig:local_graph}) and the resultant DAG will have at most $n(n-1)/2$ edges, where $n = |\Phi|$. The $local$ graph is acyclic, because the nodes are sorted by their respective evaluations in a non-increasing order and hence edges with negative weights will not be added thereby eliminating any bidirectional edges. The DAG for a demonstration imposes a partial order over all specifications. For any 2 specifications $\phix{i}$ and $\phix{j}$, $\phix{i} \succeq \phix{j}$ if $\rho(\phix{i}) \geq \rho(\phix{j})$ and so, an edge is created from $\phix{i}$ to $\phix{j}$ with weight $\rho(\phix{i}) - \rho(\phix{j})$ subject to the threshold $\epsilon$.

% Local graph figure

\textit{Complexity Analysis:} In general, given $n$ specifications and a set of algebraic operators (e.g., $op = \{>, =\}$), the number of different orderings is: $n! \cdot [|op|^{n-1} - 1] + 1$. In our case, $|op|=2$ since the operator $<$ in an ordering is equivalent to a permutation of the ordering using $>$, i.e., $a<b \equiv b>a$. By making use of directed graphs, we can eliminate the factorial component (refer to our supplemental document \citep{peglearn_supp} for proofs), but this still results in an exponential-time search algorithm. To overcome this, in our algorithm, we eliminate cycles by building a DAG for each of the $m$ demonstrations. Depending on the data structure used, the complexity of building a DAG is linear when using adjacency lists and quadratic when using adjacency matrix to represent the graph. The total complexity is thus $\mathcal{O}(m n^2)$ in the worst case (using matrix representation).

\subsection{Aggregation of \textit{local} graphs}
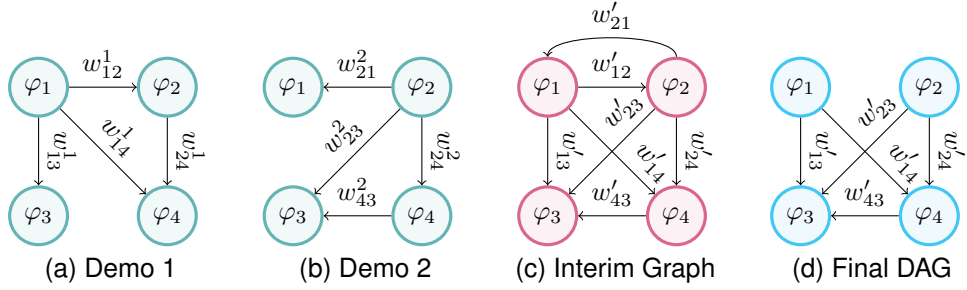
\begin{figure}[tb]
\centering
\subfloat[Demo 1\label{fig:global_step1}]{
	\begin{tikzpicture}[mypic]
		%Nodes
		\node[roundnode](phi1){$\varphi_1$};
		\node[roundnode](phi2)[right=of phi1]{$\varphi_2$};
		\node[roundnode](phi3)[below=of phi1]{$\varphi_3$};
		\node[roundnode](phi4)[below=of phi2]{$\varphi_4$};
		
		%Lines
		%From 1 to 2, 3 and 4
		\draw[->] (phi1.east) -- node[midway, above, sloped] {$w_{12}^1$} (phi2.west);
		\draw[->] (phi1.south) -- node[midway, above, sloped] {$w_{13}^1$} (phi3.north);
		\draw[->] (phi1) -- node[midway, above, sloped] {$w_{14}^1$} (phi4);
		
		% From 2 to 4
		\draw[->] (phi2.south) -- node[midway, above, sloped] {$w_{24}^1$} (phi4.north);
	\end{tikzpicture}
} \hfil
\subfloat[Demo 2]{
	\begin{tikzpicture}[mypic]
		%Nodes
		\node[roundnode](phi1){$\varphi_1$};
		\node[roundnode](phi2)[right=of phi1]{$\varphi_2$};
		\node[roundnode](phi3)[below=of phi1]{$\varphi_3$};
		\node[roundnode](phi4)[below=of phi2]{$\varphi_4$};
		
		%Lines
		% From 2 to 1, 3 and 4
		\draw[->] (phi2.west) -- node[midway, above, sloped] {$w_{21}^2$} (phi1.east);
		\draw[->] (phi2) -- node[midway, above, sloped] {$w_{23}^2$} (phi3);
		\draw[->] (phi2.south) -- node[midway, above, sloped] {$w_{24}^2$} (phi4.north);
		
		% From 4 to 3
		\draw[->] (phi4.west) -- node[midway, above, sloped] {$w_{43}^2$} (phi3.east);
		
	\end{tikzpicture}
} \hfil
\subfloat[Interim Graph \label{fig:interim_graph}]{
	\begin{tikzpicture}[mypic,
	roundnode/.style={circle, draw=purple!60, fill=purple!5, very thick, minimum size=7mm},
	]
		%Nodes
		\node[roundnode](phi1){$\varphi_1$};
		\node[roundnode](phi2)[right=of phi1]{$\varphi_2$};
		\node[roundnode](phi3)[below=of phi1]{$\varphi_3$};
		\node[roundnode](phi4)[below=of phi2]{$\varphi_4$};
		
		%Lines
		% From 1 to 2, 3 and 4
		\draw[->] (phi1.east) -- node[midway, above, sloped] {$w'_{12}$} (phi2.west);
		\draw[->] (phi1.south) -- node[midway, above, sloped] {$w'_{13}$} (phi3.north);
		\draw[->] (phi1) -- node[midway, above, sloped, pos=0.8] {$w'_{14}$} (phi4);
		
		% From 2 to 1, 3 and 4
%		\draw[->] (phi2.north) -- node[midway, above, sloped, looseness=1.5] {$w_{21}$} (phi1.north);
		\draw[->] (phi2) to [out=north, in=north, looseness=0.5] node[midway, above, sloped] {$w'_{21}$} (phi1);
		\draw[->] (phi2) -- node[midway, above, sloped, pos=0.2] {$w'_{23}$} (phi3);
		\draw[->] (phi2.south) -- node[midway, above, sloped] {$w'_{24}$} (phi4.north);
		
		% From 4 to 3
		\draw[->] (phi4.west) -- node[midway, above, sloped] {$w'_{43}$} (phi3.east);
	\end{tikzpicture}
} \hfil
\subfloat[Final DAG]{
	\begin{tikzpicture}[mypic,
	roundnode/.style={circle, draw=cyan!60, fill=cyan!5, very thick, minimum size=7mm}
	]
		%Nodes
		\node[roundnode](phi1){$\varphi_1$};
		\node[roundnode](phi2)[right=of phi1]{$\varphi_2$};
		\node[roundnode](phi3)[below=of phi1]{$\varphi_3$};
		\node[roundnode](phi4)[below=of phi2]{$\varphi_4$};
		
		%Lines
		% From 1 to 3 and 4
		\draw[->] (phi1.south) -- node[midway, above, sloped] {$w_{13}'$} (phi3.north);
		\draw[->] (phi1) -- node[midway, above, sloped, pos=0.8] {$w_{14}'$} (phi4);
		
		% From 2 to 3 and 4
		\draw[->] (phi2) -- node[midway, above, sloped, pos=0.2] {$w_{23}'$} (phi3);
		\draw[->] (phi2.south) -- node[midway, above, sloped] {$w_{24}'$} (phi4.north);
		
		% From 4 to 3
		\draw[->] (phi4.west) -- node[midway, above, sloped] {$w_{43}'$} (phi3.east);
	\end{tikzpicture}
} \hfil
\caption{Example $global$ graph from 2 demonstrations.}
\label{fig:global_graph}
\end{figure}

Once the $local$ graphs for each demonstration have been generated, they need to be combined into a single DAG to be used directly in the LfD-STL framework \citep{puranic_corl2020, puranic_ral2021}. We now propose \autoref{alg:peglearn_global} to aggregate all $local$ graphs into a single DAG. Line 2 generates the $local$ graphs via \autoref{alg:peglearn_local} and stores them in a dataset $\mathcal{G}$. For every directed edge between any pairs of vertices $u$ and $v$, the mean of the weights on corresponding edges across all graphs in $\mathcal{G}$ is computed (line 3 of \autoref{alg:peglearn_global}). For example, consider the $local$ graphs of 2 sample demonstrations shown in \autoref{fig:global_graph}. By averaging the edge weights of the graphs of the 2 demonstrations, we get the intermediate weighted directed graph shown in \autoref{fig:interim_graph}. This is not necessarily acyclic since there is a cycle between the nodes of $\varphi_1$ and $\varphi_2$.
In this figure, each $w_{ij}' = (w_{ij}^1 + w_{ij}^2) / 2$. This intermediate graph needs to be further reduced to a weighted DAG, i.e., by eliminating any cycles/loops.

\begin{algorithm}[htbp]
% \SetAlgoLined
\DontPrintSemicolon
\SetKwInput{Input}{Input}
\Input{$D$ := set of $m$ demonstrations; $\Phi$ := set of $n$ specifications; $\epsilon$: threshold (tunable)
} 
\KwResult{Constructs global Performance-Graph}
\BlankLine

\Begin{
$\mathcal{G} \leftarrow \bigcup_{i=1}^m G_i$ \tcp*[r]{via \autoref{alg:peglearn_local}}

$G' \leftarrow \frac{1}{|\mathcal{G}|} \sum_{G \in \mathcal{G}} G$ \tcp*[r]{Edge-wise mean}

%% REDUCED GRAPH FUNCTION %%
\tcp{Extracts edge-weighted DAG from the raw Performance-Graph}
$\mathbf{G} \leftarrow \mathbf{0_{n \times n}}$ \tcp*[r]{zero matrix}
\For{$i=1$ to $n$}{
	\For{$j=1$ to $n$}{
		$\mathbf{G}[i, j] \leftarrow max(0, G'[i, j] - G'[j, i])$ \;
		\lIf{$\mathbf{G}[i, j] < \epsilon$} {$\mathbf{G}[i, j] \leftarrow 0$}
	}
}
\Return $\mathbf{G}$ \;
}
\caption{PeGLearn: Generating the global DAG from all demonstrations. \label{alg:peglearn_global}}
\end{algorithm}

\subsection{Conversion/Reduction to weighted DAG}
\label{sec:theory}
Note that there can only be at most 2 edges between any pair of vertices since the outgoing (and similarly, incoming) edges are averaged into a single edge. In order to reduce this graph to a global DAG, we systematically eliminate edges by first computing the difference between the outgoing and incoming edge and then checking if it is above a certain threshold to add an edge in the direction of positive difference (note that if the difference is negative, the edge can be simply reversed). In other words, for any 2 nodes, $u$ and $v$, if $(w(u, v) - w(v, u)) \ge \epsilon$, then $e(u, v)$ is retained with new weight $w(u, v) - w(v, u)$, while $e(v, u)$ is removed or discarded since it gets absorbed by the retained edge. The threshold $\epsilon$ again acts as a high-pass filter. As we can observe in the case of bidirectional edges, one of the edges will be ``consumed'' by the other or both will be discarded if they are similar. This conversion procedure is shown by lines 5--8 in \autoref{alg:peglearn_global}. Thus, all cycles/loops are eliminated, resulting in a weighted DAG that can be directly used to rank the demonstrations and compute rewards for RL tasks as performed in the LfD-STL framework. To show that our DAG-learning method indeed preserves the performance ranking over demonstrations, we first define a partial ordering over demonstrations: for any 2 demonstrations $\demo_1$ and $\demo_2$, the partial order $\demo_1 \preceq \demo_2$ is defined when $\rho_{1i} \leq \rho_{2i}, \forall i \in \{1, \cdots, n\}$. Thus, we say that $\demo_2$ is better or at least as good as $\demo_1$. Then, by making use of \autoref{lemma}, we arrive at \autoref{theorem} that addresses the problem definition (all proofs are in \citep{peglearn_supp}).
\begin{lemma}
For a DAG, the weights associated with the nodes computed via \eqref{eqn:dag_weights} are non-negative.
\label{lemma}
\end{lemma}

\begin{thm}
For any two demonstrations $\demo_1$ and $\demo_2$ in an environment, the partial ordering $\demo_1 \preceq \demo_2$ is preserved by PeGLearn. 
\label{theorem}
\end{thm}
The global DAG imposes a partial order of specifications. For any 2 specifications $\varphi_i$ and $\varphi_j$, the partial order $\varphi_i \succeq \varphi_j$ is defined when $\bar{\rho}(\phix{i}) \geq \bar{\rho}(\phix{j})$, where $\bar{\rho}(\varphi)$ is the mean of $\rho(\varphi)$ across all demonstrations. The $global$ graph thus uses a holistic approach to explain the overall performance of demonstrations and could provide an intuitive representation for non-expert users to teach agents to do tasks as well as understand the policies the agent is learning.

\section{Experiments}
\label{sec:exps}

\subsection{Comparison with baseline}
For comparison with user-defined DAG in the LfD-STL baseline \citep{puranic_corl2020, puranic_ral2021}, we evaluated our method on the same discrete-world and 2D autonomous driving domains using the same demonstrations ($m=8$) and specifications. We show the results for the 2D driving scenario in \autoref{fig:dubins2d}. The STL specifications correspond to (i) reaching the goal $\varphi_1$, (ii) avoiding the hindrance/obstacle regions $\varphi_2$, (iii) always staying within the workspace region $\varphi_3$, and (iv) reaching the goal as fast as possible $\varphi_4$. This resembles a real-world scenario wherein, one of the challenging problems in autonomous driving is overtaking moving or stationary/parked vehicles on road-sides (e.g., urban and residential driving). The scenario presented here is a high-level abstraction where the purple square is a parked car and the yellow square is the goal state of the ego car after overtaking the parked car. The light-yellow shaded region are the dimensions of the road/lane and the task for the ego car is to navigate around the parked car to the goal state without exiting the lane. From these figures, we can observe that the rewards using the inferred DAG are consistent with the specifications, i.e., rewards are aligned with entity locations. In the discrete-world settings, we were able to learn similar graphs and rewards, and hence {\em same policies} as the baseline. This shows that our proposed method is a significant improvement over the prior work since it eliminates the burden of the user to define graphs, while also using at least {\em 4 times} fewer demonstrations than IRL-based methods {\citep{ziebart_maximumentropy, mce_irl_ziebart}. 

\begin{figure*}[tb]
\centering
\subfloat[Learned DAG]{
	\begin{tikzpicture}[mypic]
		%Nodes
		\node[roundnode](phi1){$\varphi_1$};
		\node[roundnode](phi2)[right=of phi1]{$\varphi_2$};
		\node[roundnode](phi3)[below=of phi1]{$\varphi_3$};
		\node[roundnode](phi4)[below=of phi2]{$\varphi_4$};
		
		%Lines

		% From 1 to 2, 3 and 4
		\draw[->] (phi1.east) -- (phi2.west);
		\draw[->] (phi1.south) -- (phi3.north);
		\draw[->] (phi1) -- (phi4);
		
		% From 2 to 4
		\draw[->] (phi2.south) -- (phi4.north);
		
		% From 3 to 4
		\draw[->] (phi3.east) -- (phi4.west);
		
	\end{tikzpicture}
	}\hfil
\subfloat[Driving layout]{\includegraphics[width=1.4in, height=1.2in]{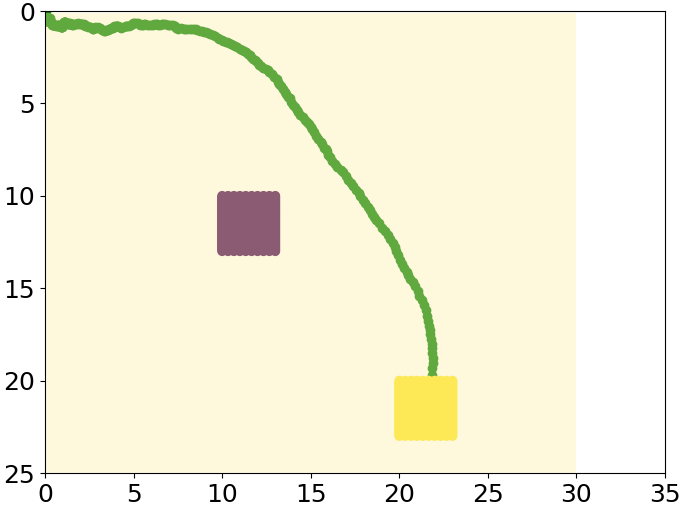}} \hfil
\subfloat[Baseline rewards]{\includegraphics[width=1.4in, height=1.2in]{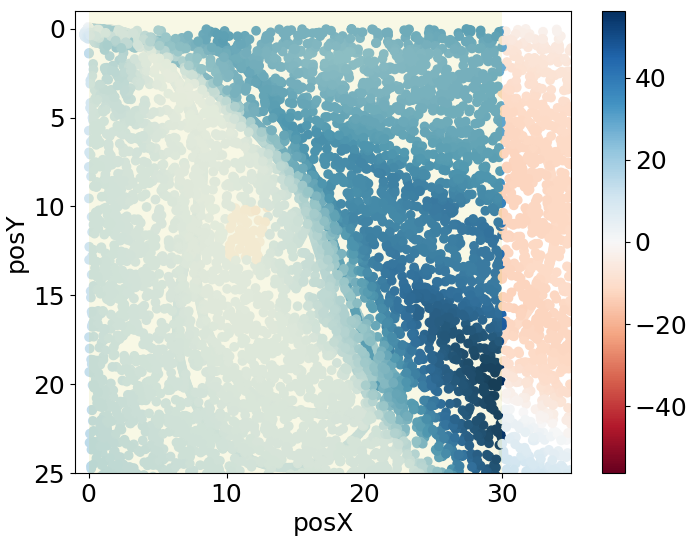}} \hfil
\subfloat[PeGLearn rewards]{\includegraphics[width=1.4in, height=1.2in]{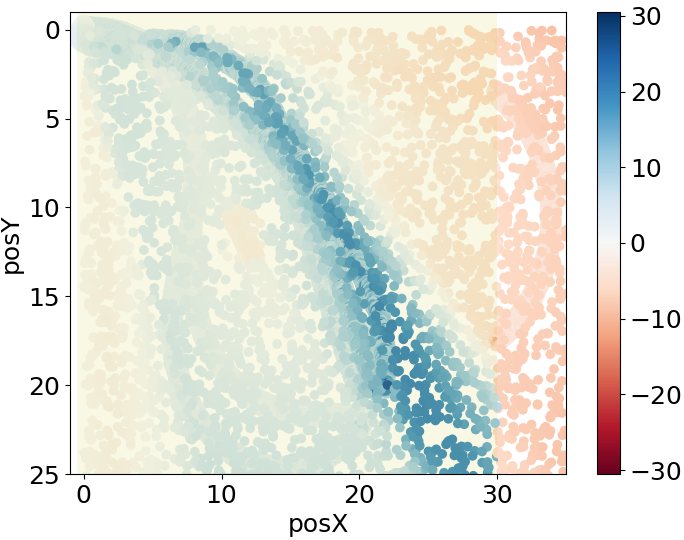}}
\caption{Results for the 2-D autonomous driving simulator. Baseline rewards are from user-defined DAG.}
\label{fig:dubins2d}
\end{figure*}
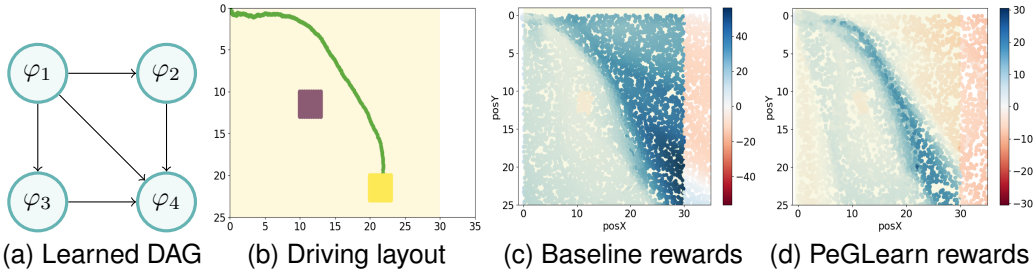

\subsection{Industrial Mobile Robot Navigation}
\begin{figure*}[htbp]
\centering
\subfloat[Environment setup]{\includegraphics[width=1.4in, height=1.2in]{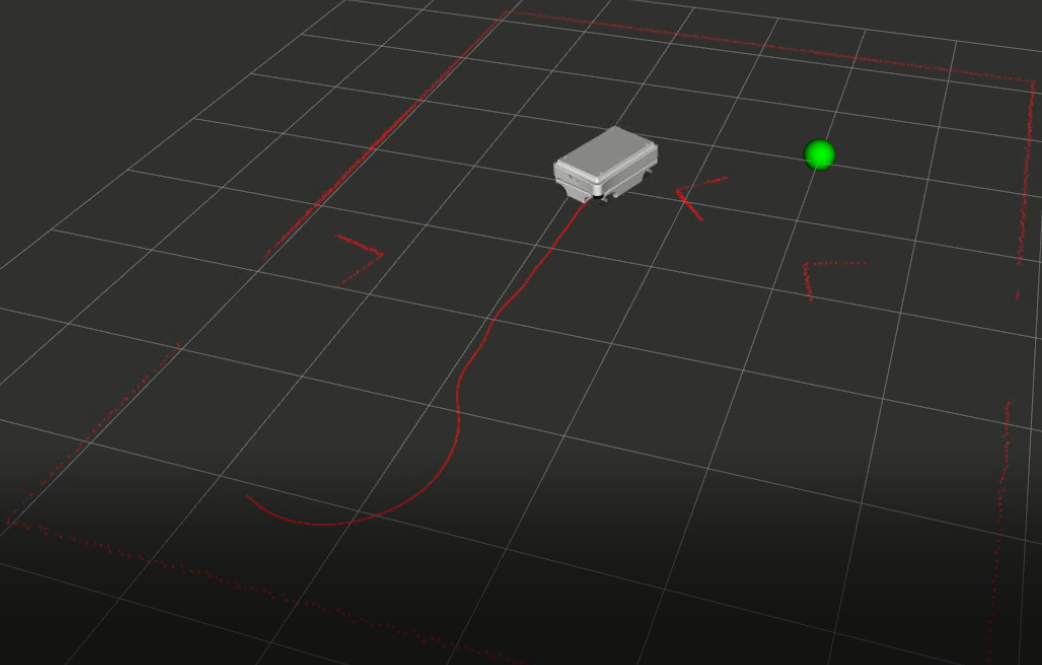}}\hfil
\subfloat[Learned DAG]{\includegraphics[height=1.2in]{mir100/peglearn}}

\subfloat[Expert rewards]{\includegraphics[width=1.4in, height=1.2in]{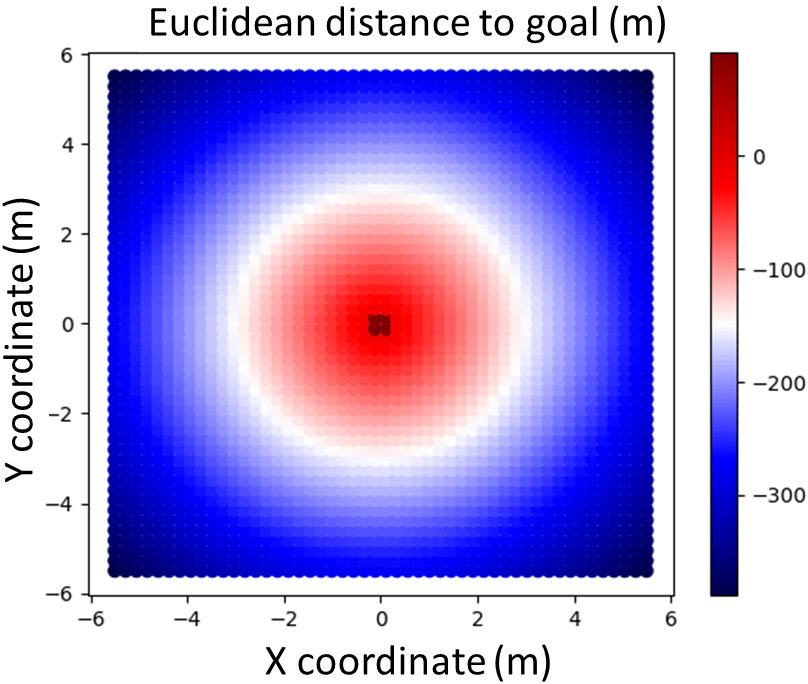}}\hfil
\subfloat[PeGLearn rewards]{\includegraphics[width=1.4in, height=1.2in]{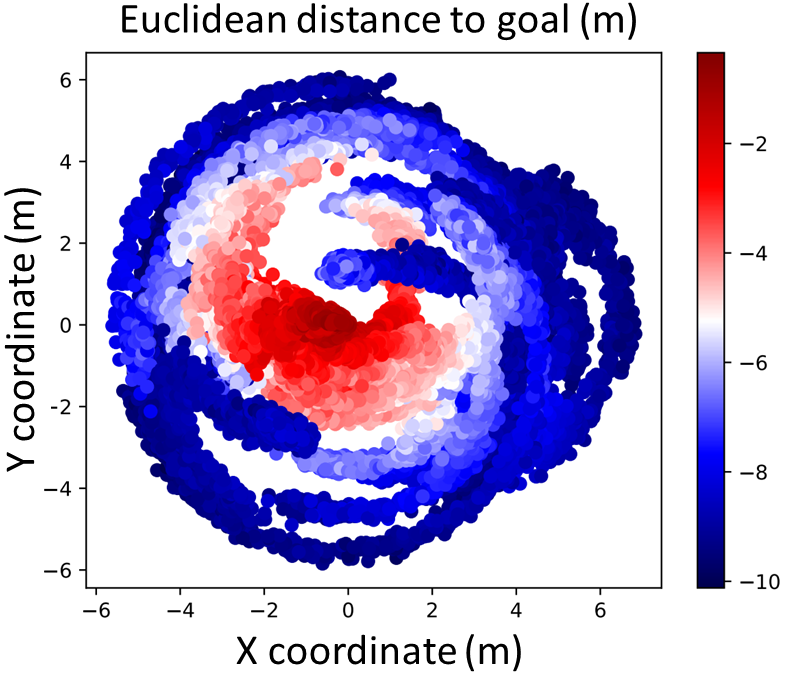}}
\caption{Results for the MiR100 navigation environment. (a) Obstacles and boundary walls are shown in red point clouds; green sphere is the goal.}
\label{fig:mir100}
\end{figure*}
In this setup, we consider a high-fidelity simulator \citep{lucchi2020robo} based on the mobile robot {\em MiR100} that is widely used in today's industries\footnote{This paper is accompanied by a multimedia (video) for the experiments. \label{fn:exp}}. In this environment, the robot is tasked with navigating to a goal location while avoiding 3 obstacles (\autoref{fig:mir100}). However, the locations of the robot, goal and 3 obstacles are randomly initialized for every episode. This presents a major challenge for LfD algorithms since the demonstrations collected on this environment are unique to a particular configuration of the entity locations (i.e., no two demonstrations are the same). The 20-dimensional state-space of the robot, indexed by timestep $t$ consists of the position of the goal in the robot's frame in polar coordinates (radial distance $p_t$ and orientation $\theta_t$), linear ($v_t$) and angular ($w_t$) velocities, and 16 readings from the robot's Lidar for detecting obstacles ($x_t^i, i \in [1,16]$). We obtained 30 demonstrations, of which 15 were incomplete (i.e., collided with obstacle or failed to reach the goal in time) by training an RL agent on an expert reward function and recording trajectories at different training intervals. The specifications governing this environment are:

\begin{enumerate}[leftmargin=1em,nosep]
\item Eventually reach the goal: \\ $\varphi_g:= \ev (p_t \leq \delta)$, where $\delta$ is a small threshold set by the environment to determine if the robot is sufficiently close to the goal to terminate the episode.
\item Always maintain linear and angular velocity limits: \\ $\varphi_v:= \alw (v_{min} \leq v_t \leq v_{max})$, and $\varphi_w:= \alw (w_{min} \leq w_t \leq w_{max})$. The limits conform to the robot's capabilities.
\item Always avoid obstacles: $\varphi_s:= \alw (\bigwedge_{i=1}^{16} (x_t^i > 0))$, where $x_i^t$ is the distance from an obstacle as measured by Lidar $i$. 
\item Reach the goal in a reasonable time: \\ $\varphi_t:= \ev (T_{min} \leq t \leq T_{max})$, where the time limits are obtained from the average lengths of $good$ demonstrations observed in this environment.
\end{enumerate}

Once the graph is extracted via PeGLearn, the rewards are propagated to the observed states and modeled with a neural network via the method described in \citep{puranic_ral2021}. We used a 2-hidden layer neural network with 512 nodes in each layer for reward approximation. The DAG and rewards inferred are shown in \autoref{fig:mir100}. We can observe that the rewards are semantically consistent with the specifications, in that, rewards increase as the robot moves towards the goal (center of figure) in a radial manner. To compare the policies learned with our method and expert-designed dense rewards, we independently trained two RL agents via: (i) PPO (on-policy, stochastic actions) \citep{ppo_schulman} and (ii) D4PG (off-policy, deterministic actions) \citep{d4pg}, on each of the reward functions and evaluated the policies over 100 trials.

Despite being presented with only 50\% successful demonstrations, our method achieved a success rate of (i) $\mathbf{79\%}$ compared to 81\% for expert rewards when using PPO, with {\em 29\%} improvement over demonstrations, and (ii) $\mathbf{90\%}$ compared to 93\% for expert rewards when using D4PG, with {\em 40\%} improvement over demonstrations. This indicates that our method is capable of producing expert-like behaviors. Furthermore, this particular environment \citep{lucchi2020robo} has shown to be readily transferable to real-robots without any modifications (i.e., the sim2real transfer gap is almost nil), which also indicates the real-world applicability of our framework.

\subsection{CARLA Driving Simulator}
We evaluated our method on a realistic driving simulator, CARLA \citep{carla}, on highway and urban driving scenarios. A demonstrator controls the (ego) car via an analog controller. The states of the car provided by the environment are: lateral distance and heading error between the ego vehicle to the target lane center line (in \texttt{meter} and \texttt{rad}), ego vehicle's speed (in \texttt{meters per second}), and (\texttt{Boolean}) indicator of whether there is a front vehicle within a safety margin. Based on this information, we formulated 3 STL specifications that are similar to the 2D driving and mobile navigation experiments, which informally, correspond to (i) keeping close to the lane center $\varphi_1$, (ii) maintaining speed limits $\varphi_2$ and (iii) maintaining safe distance to any lead vehicle $\varphi_3$; formal definitions are provided in \citep{peglearn_supp}. For this scenario, we recorded 15 demonstrations from one of the authors of this paper via an analog steering and pedal controller. These 15 videos were split uniformly across 3 batches where, the 5 videos in each batch showed a common behavior, but the behaviors were different across the batches. All videos were exclusive to their respective batches, i.e., no video was used in more than 1 batch, and each video was 30 seconds long on average.

\mypara{User Study}
The recorded driving videos were used to perform a user study to determine if users would rate the driving behavior similarly, thereby providing evidence that the graphs generated using PeGLearn produce accurate ranking of specifications\footref{fn:exp}. Using the Amazon Mechanical Turk (AMT) platform, we created a survey of the 3 batches representing the split of the 15 videos. Each participant of a batch was shown the corresponding 5 driving videos and was tasked with providing $Likert$ ratings for: (i) performance of demonstrator on each of the 3 task specifications described in natural language and (ii) ranking of specifications based on overall driving behaviors. We recruited 146 human participants via AMT service, each with an approval rating over $98\%$. The average driving experience of the participants was 22.4 years. This user study was approved by the {\em University of Southern California Institution Review Board}
%\footnote{Reference email: irb@usc.edu}
on {\em January 10, 2022}. Additional information about this survey and participant demographics are provided in the supplemental document \citep{peglearn_supp}. The goal of this study was to first ensure that the PeGLearn orderings were similar to expert orderings and not a random coincidence. The total number of possible orderings for the 3 specifications is 27 $(=3^3)$, so for each video and participant, we also generated an ordering randomly and uniformly chosen from the space of 27 orderings. Based on this, we formulate the hypothesis as: 

\begin{hypothesis}
The similarity between human expert and PeGLearn orderings is significantly higher than that of a random ordering.
\end{hypothesis}

Secondly, we wanted to investigate the similarity between existing clustering techniques such as K-Means \citep{bishopPRML, murphyML} and our algorithm, with expert rankings. The problem of finding weights for specifications resembles anomaly detection where the \textit{bad} demonstrations are outliers and methods such as clustering, classification or combination of both can be employed \citep{kmeans_svm}. Additionally, the weights for specifications that we seek to learn, indicate the importance of the specifications, which is analogous to importance/rank of features in classification tasks \citep{feature_rank}. Hence, we use K-Means combined with SVM \citep{bishopPRML, murphyML} for comparison purposes, which we now refer to as Km+SVM. Clustering is first employed to extract the clusters of demonstrations based on their corresponding ratings. Note that the set of demonstrations can contain all {\em good} (ones with positive ratings for all specifications), all {\em bad} (ones with negative ratings for any specification) or a mixture of both these types. Based on the inertia of the k-means clustering for this data, we found that $k=2$ was optimal for each batch and provided the best fit. Then, SVM was used to classify the cluster centroids and extract the weights. The magnitude of the weights indicate the relative importance/ranking of each feature (specification) \citep{feature_rank}. So we ranked the weights to compare with PeGLearn rankings. Therefore, we formulate our second hypothesis as follows:

\begin{hypothesis}
The similarity between human expert and PeGLearn orderings is significantly higher than that of K-means+SVM.
\end{hypothesis}

\mypara{Analysis}
We first obtain the ratings and hence specification-orderings from all sources: participants, PeGLearn, Km+SVM and uniform-random algorithm. We then compute the \hamm distance \citep{hamming_distance} between the human expert orderings and orderings from (i) PeGLearn, (ii) Km+SVM, and (iii) uniform-random. The reason being that the \hamm distance between any two sequences of equal lengths measures the number of element-wise disagreements or mismatches, and hence gives an estimate of how close any two orderings are. The distance is a value in $[0, 1]$ with 0 representing same sequences and 1 indicating completely different sequences. To perform statistical analysis, we introduce a few notations for convenience, as follows: (i) PH for PeGLearn--human \hamm distance or error, (ii) KH for Km+SVM--human \hamm error, and (iii) RH for uniform-random--human \hamm error. We concatenate these errors under the name ``Score'' for analysis purposes. Note that lower the \hamm distance or ``Score'', the more similar are the two orderings. A two-way ANOVA was conducted to examine the effects of agent type (i.e., \{PH, KH, RH\}) and batch number (i.e., \{1, 2, 3\}) on the ``Score''. There was no statistically significant interaction between agent type and batch number for ``Score'', $F(4, 429) = 1.605, p = .172$, partial $\eta^2 = .015$. Therefore, an analysis of the main effect of agent type was performed, which indicated there was a statistically significant main effect of agent type, $F(2, 429) = 34.558, p < .001$, partial $\eta^2 = .139$. All pairwise comparisons were run, where reported $95\%$ confidence intervals and p-values are Bonferroni-adjusted. The marginal means for ``Score'' were .475 (SE = .022) for PH, .724 (SE = .022) for KH and .660 (SE = .022) for RH. The ``Score'' means between RH and PH were found to be statistically significant $p < .001$, showing support for \textbf{H1}. Similarly, the ``Score'' means for PH and KH were statistically significant $p < .001$ and the mean for KH was higher than PH, supporting \textbf{H2}. Lastly, there was no statistically significant main effect of batch number on ``Score'', $F(2, 429) = .172, p = .842$, partial $\eta^2 = .001$.

\mypara{Comparison with Km+SVM} K-means typically has a complexity of $\mathcal{O}(mknt)$ where $m$ is the number of data points (i.e., demonstrations), $k$ is the number of components/clusters, $n$ is the dimension of data (i.e., number of specifications) and $t$ is the number of iterations. Linear SVM follows linear complexity in $m$ and so the combination of Km+SVM is still $\mathcal{O}(mknt)$. Since there are $k=2$ components in our formulation, the $k$ is treated as a constant and the complexity is just $\mathcal{O}(mnt)$. Our algorithm on the other hand has a complexity of $\mathcal{O}(mn^2)$ when using matrices to represent graphs. This shows that our algorithm not only performs better than clustering methods, but is also more efficient because generally, the number of specifications is much smaller than the number of iterations to converge ($n \ll t$). {\em All the experiments and results show that our method can not only learn accurate rewards similar to the way humans perceive them, but it does so with a limited number of even imperfect data}. Similarly, we also performed experiments on a {\em real-world} surgical robot dataset, JIGSAWS, to demonstrate how human $Likert$ ratings can be used to learn DAGs \citep{peglearn_supp}. Additionally, together with the LfD-STL framework, we are able to learn temporal-based rewards, even in continuous and high-dimensional spaces with just a handful of demonstrations.

\section{Related Work}
\label{sec:rel_work}

% LfD in general
Learning-from-demonstrations (LfD) has been extensively explored to obtain control policies for robots \citep{lfd_schaal1, lfd_schaal2} using methods such as behavioral cloning \citep{bco_stone} and inverse reinforcement learning (IRL) \citep{Ng_russell, ziebart_maximumentropy, abbeel_ng}. 

% Learning with Temporal Logics
Aside from our prior works, there have been many other methods involving temporal logics with demonstrations. A counterexample-guided approach using probabilistic computation tree logics for safety-aware apprenticeship learning is proposed in \citep{wenchao_cav}, which makes use of DAGs to search for counterexamples. The authors in \citep{cho_mpc_stl} utilize model-predictive control to design a controller that imitates demonstrators while deciding the trade-offs among STL rules. Similar to our prior works, they assume that the priorities or ordering among the STL specifications are provided beforehand. An active learning approach has been explored in \citep{memarian_cdc2020}, in which the rewards are learned by augmenting the state space of a Markov Decision Process (MDP) with an automaton, represented by directed graphs. An alternative approach to characterize the expressivity of Markovian rewards has been proposed in \citep{abel2021expressivity}, which could provide interpretations for rewards in terms of task descriptions. However, our LfD-STL framework offers several critical advantages over IRL and the methods of \citep{abel2021expressivity}, such as: employing non-Markovian rewards via task-based temporal logics and empirically having a significant reduction in sample and computation complexity. Additionally, the methods in \citep{abel2021expressivity} have been explored for deterministic systems, while LfD-STL can also generalize to stochastic dynamics and continuous spaces as seen in our experiments.

% Explainable RL
Causal influence diagrams of MDPs via DAGs for explainable online-RL have been recently investigated \citep{everitt_reward_2021}. Another related work by the authors of \citep{Madumal_Miller_Sonenberg_Vetere_2020} make use of causal models to extract causal explanations of the behavior of model-free RL agents. They represent action influence models via DAGs in which the nodes are random variables representing the agent's world and edges correspond to the actions. These works are mainly focused on explainability in forward RL (i.e., when rewards are already known), while our method is mainly focused on generating intuitive representations of behaviors and rewards to be used later in forward RL.

% Explainable Rewards
In the area of reward explanations for RL tasks, the method proposed in \citep{xrl_reward_decompose} decomposes rewards into several components, based on which, the RL agent's action preferences can be explained and can help in finding bugs in rewards. Another work pertaining to IRL \citep{score_irl} uses expert-scored trajectories to learn a reward function. This work, which builds on standard IRL, typically relies on a large dataset containing several hundreds of nearly-optimal demonstrations and hence generating scores for each of them. By \autoref{theorem}, our method can also overcome any rank-conflicts arising out of myopic trajectory preferences \citep{biyik_preference}. The authors in \citep{sanneman2022} have investigated the reward-explanation problem in the context of human-robot teams wherein the robot, via interactions, learns the reward that is known to the human. They propose 2 categories of reward explanations: (i) feature-space: where rewards are explained through individual features comprising the reward function and their relative weights, and (ii) policy-space: where demonstrations of actions under a reward function are used to explain the rewards. Our work can be regarded as a combination of these categories since it uses specifications as features along with inferred weights, and demonstrations.

\section{Conclusion}
In this work, we proposed a novel algorithm, {\em PeGLearn}, to capture the performance and provide intuitive holistic representations of demonstrations in the form of graphs. We showed, through challenging experiments in robotic domains, that the inferred graphs could be directly applied to the existing LfD-STL framework to extract rewards and robust control policies via RL, with a limited number of even imperfect demonstrations. The user study conducted showed that our graph-based method produced more accurate results than (un)supervised algorithms in terms of similarities with human ratings. We believe our work is a step in the direction of developing interpretable and explainable learning systems with formal guarantees, which is one of the prominent challenges today \citep{naesm_humanAI,gazit_fmhri2021, abel2021expressivity}. Using intuitive structures such as DAGs to represent rewards and trajectories would provide insights into the learning aspects of RL agents, as to the quality of behaviors they are learning and can be used alongside/integrated with works in explainable AI \citep{gombolay_xai, sanneman2022}. In the future, we will also investigate the use of such graphs during RL as feedback to improve policies and provide performance guarantees.

\bibliography{cps,learning,surveys}

\begin{thebibliography}{52}
\providecommand{\natexlab}[1]{#1}
\providecommand{\url}[1]{\texttt{#1}}
\expandafter\ifx\csname urlstyle\endcsname\relax
  \providecommand{\doi}[1]{doi: #1}\else
  \providecommand{\doi}{doi: \begingroup \urlstyle{rm}\Url}\fi

\bibitem[{National Academies of Sciences, Engineering, and
  Medicine}(2022)]{naesm_humanAI}
{National Academies of Sciences, Engineering, and Medicine}.
\newblock \emph{Human-AI Teaming: State-of-the-Art and Research Needs}.
\newblock The National Academies Press, 2022.
\newblock ISBN 978-0-309-27017-5.
\newblock \doi{10.17226/26355}.

\bibitem[Kress-Gazit et~al.(2021)Kress-Gazit, Eder, Hoffman, Admoni, Argall,
  Ehlers, Heckman, Jansen, Knepper, K\v{r}et\'{\i}nsk\'{y}, Levy-Tzedek, Li,
  Murphey, Riek, and Sadigh]{gazit_fmhri2021}
H.~Kress-Gazit, K.~Eder, G.~Hoffman, H.~Admoni, B.~Argall, R.~Ehlers,
  C.~Heckman, N.~Jansen, R.~Knepper, J.~K\v{r}et\'{\i}nsk\'{y}, S.~Levy-Tzedek,
  J.~Li, T.~Murphey, L.~Riek, and D.~Sadigh.
\newblock Formalizing and guaranteeing human-robot interaction.
\newblock \emph{Commun. ACM}, 2021.

\bibitem[Paleja et~al.(2021)Paleja, Ghuy, Arachchige, Jensen, and
  Gombolay]{gombolay_xai}
R.~Paleja, M.~Ghuy, N.~R. Arachchige, R.~Jensen, and M.~Gombolay.
\newblock The utility of explainable ai in ad hoc human-machine teaming.
\newblock In \emph{NeurIPS}, 2021.

\bibitem[Sanneman and Shah(2022)]{sanneman2022}
L.~Sanneman and J.~A. Shah.
\newblock An empirical study of reward explanations with human-robot
  interaction applications.
\newblock \emph{RA-L}, 2022.

\bibitem[Amodei et~al.(2016)Amodei, Olah, Steinhardt, Christiano, Schulman, and
  Mané]{amodei2016concrete}
D.~Amodei, C.~Olah, J.~Steinhardt, P.~Christiano, J.~Schulman, and D.~Mané.
\newblock Concrete problems in {AI} safety, 2016.

\bibitem[Gundana and Kress{-}Gazit(2021)]{gundana_stl}
D.~Gundana and H.~Kress{-}Gazit.
\newblock Event-based signal temporal logic synthesis for single and
  multi-robot tasks.
\newblock \emph{{IEEE} RA-L}, 2021.

\bibitem[Li et~al.(2017)Li, Vasile, and Belta]{li_reinforcement_2017}
X.~Li, C.~Vasile, and C.~Belta.
\newblock Reinforcement learning with temporal logic rewards.
\newblock In \emph{IROS}, 2017.

\bibitem[Li et~al.(2018)Li, Ma, and Belta]{belta_lfd}
X.~Li, Y.~Ma, and C.~Belta.
\newblock Automata guided reinforcement learning with demonstrations, 2018.

\bibitem[Aksaray et~al.(2016)Aksaray, Jones, Kong, Schwager, and
  Belta]{aksaray_q-learning_2016}
D.~Aksaray, A.~Jones, Z.~Kong, M.~Schwager, and C.~Belta.
\newblock Q-{{Learning}} for robust satisfaction of signal temporal logic
  specifications.
\newblock In \emph{Conference on Decision and Control (CDC)}, 2016.

\bibitem[Wen et~al.(2015)Wen, Ehlers, and Topcu]{wen_2015}
M.~Wen, R.~Ehlers, and U.~Topcu.
\newblock Correct-by-synthesis reinforcement learning with temporal logic
  constraints.
\newblock In \emph{IROS}, 2015.

\bibitem[Maler and Nickovic(2004)]{stl_complexity}
O.~Maler and D.~Nickovic.
\newblock Monitoring temporal properties of continuous signals.
\newblock In Y.~Lakhnech and S.~Yovine, editors, \emph{Formal Techniques,
  Modelling and Analysis of Timed and Fault-Tolerant Systems}, pages 152--166,
  Berlin, Heidelberg, 2004. Springer Berlin Heidelberg.
\newblock ISBN 978-3-540-30206-3.

\bibitem[Puranic et~al.(2020)Puranic, Deshmukh, and
  Nikolaidis]{puranic_corl2020}
A.~G. Puranic, J.~V. Deshmukh, and S.~Nikolaidis.
\newblock Learning from demonstrations using signal temporal logic.
\newblock In \emph{CoRL}, 2020.

\bibitem[Puranic et~al.(2021)Puranic, Deshmukh, and
  Nikolaidis]{puranic_ral2021}
A.~G. Puranic, J.~V. Deshmukh, and S.~Nikolaidis.
\newblock Learning from demonstrations using signal temporal logic.
\newblock In \emph{IEEE Robotics and Automation Letters (RA-L)}, 2021.

\bibitem[Ziebart(2010)]{mce_irl_ziebart}
B.~D. Ziebart.
\newblock \emph{Modeling Purposeful Adaptive Behavior with the Principle of
  Maximum Causal Entropy}.
\newblock PhD thesis, Carnegie Mellon University, USA, 2010.

\bibitem[Suay et~al.(2016)Suay, Brys, Taylor, and Chernova]{suay_aamas16}
H.~B. Suay, T.~Brys, M.~E. Taylor, and S.~Chernova.
\newblock Learning from demonstration for shaping through inverse reinforcement
  learning.
\newblock In \emph{AAMAS}, 2016.

\bibitem[Fu et~al.(2018)Fu, Luo, and Levine]{adv_irl}
J.~Fu, K.~Luo, and S.~Levine.
\newblock Learning robust rewards with adverserial inverse reinforcement
  learning.
\newblock In \emph{ICLR}, 2018.

\bibitem[Norman(2010)]{norman2010likert}
G.~Norman.
\newblock Likert scales, levels of measurement and the “laws” of
  statistics.
\newblock \emph{Advances in health sciences education}, 2010.

\bibitem[Chen et~al.(2018)Chen, Nikolaidis, Soh, Hsu, and
  Srinivasa]{chen_niko_trust2018}
M.~Chen, S.~Nikolaidis, H.~Soh, D.~Hsu, and S.~Srinivasa.
\newblock Planning with trust for human-robot collaboration.
\newblock In \emph{HRI}, 2018.

\bibitem[Puranic et~al.(2022)Puranic, Deshmukh, and Nikolaidis]{peglearn_supp}
A.~G. Puranic, J.~V. Deshmukh, and S.~Nikolaidis.
\newblock Learning performance graphs from demonstrations via task-based
  evaluations - supplemental material, 2022.
\newblock URL
  \url{https://aniruddh-puranic.info/files/performance_graph_supplemental.pdf}.

\bibitem[Ziebart et~al.(2008)Ziebart, Maas, Bagnell, and
  Dey]{ziebart_maximumentropy}
B.~D. Ziebart, A.~L. Maas, J.~A. Bagnell, and A.~K. Dey.
\newblock Maximum entropy inverse reinforcement learning.
\newblock In \emph{AAAI}, 2008.

\bibitem[Lucchi et~al.(2020)Lucchi, Zindler, M{\"u}hlbacher-Karrer, and
  Pichler]{lucchi2020robo}
M.~Lucchi, F.~Zindler, S.~M{\"u}hlbacher-Karrer, and H.~Pichler.
\newblock robo-gym--an open source toolkit for distributed deep reinforcement
  learning on real and simulated robots.
\newblock \emph{IROS}, 2020.

\bibitem[Schulman et~al.(2017)Schulman, Wolski, Dhariwal, Radford, and
  Klimov]{ppo_schulman}
J.~Schulman, F.~Wolski, P.~Dhariwal, A.~Radford, and O.~Klimov.
\newblock Proximal policy optimization algorithms, 2017.

\bibitem[Barth{-}Maron et~al.(2018)Barth{-}Maron, Hoffman, Budden, Dabney,
  Horgan, TB, Muldal, Heess, and Lillicrap]{d4pg}
G.~Barth{-}Maron, M.~W. Hoffman, D.~Budden, W.~Dabney, D.~Horgan, D.~TB,
  A.~Muldal, N.~Heess, and T.~P. Lillicrap.
\newblock Distributed distributional deterministic policy gradients.
\newblock In \emph{ICLR}, 2018.

\bibitem[Dosovitskiy et~al.(2017)Dosovitskiy, Ros, Codevilla, Lopez, and
  Koltun]{carla}
A.~Dosovitskiy, G.~Ros, F.~Codevilla, A.~Lopez, and V.~Koltun.
\newblock {CARLA}: {An} open urban driving simulator.
\newblock In \emph{CoRL}, 2017.

\bibitem[Bishop(2006)]{bishopPRML}
C.~M. Bishop.
\newblock \emph{Pattern Recognition and Machine Learning}.
\newblock Springer, 2006.

\bibitem[Murphy(2012)]{murphyML}
K.~P. Murphy.
\newblock \emph{Machine Learning: A Probabilistic Perspective}.
\newblock The MIT Press, 2012.

\bibitem[Zheng et~al.(2014)Zheng, Yoon, and Lam]{kmeans_svm}
B.~Zheng, S.~W. Yoon, and S.~S. Lam.
\newblock Breast cancer diagnosis based on feature extraction using a hybrid of
  k-means and support vector machine algorithms.
\newblock \emph{Expert Systems with Applications}, 2014.

\bibitem[Guyon et~al.(2002)Guyon, Weston, Barnhill, and Vapnik]{feature_rank}
I.~Guyon, J.~Weston, S.~Barnhill, and V.~Vapnik.
\newblock Gene selection for cancer classification using support vector
  machines.
\newblock \emph{Mach. Learn.}, 2002.

\bibitem[Hamming(1950)]{hamming_distance}
R.~W. Hamming.
\newblock Error detecting and error correcting codes.
\newblock \emph{The Bell System Technical Journal}, 29\penalty0 (2):\penalty0
  147--160, 1950.

\bibitem[Atkeson and Schaal(1997)]{lfd_schaal1}
C.~G. Atkeson and S.~Schaal.
\newblock Robot learning from demonstration.
\newblock In \emph{ICML}, 1997.

\bibitem[Schaal(1996)]{lfd_schaal2}
S.~Schaal.
\newblock Learning from demonstration.
\newblock In \emph{NIPS}, 1996.

\bibitem[Torabi et~al.(2018)Torabi, Warnell, and Stone]{bco_stone}
F.~Torabi, G.~Warnell, and P.~Stone.
\newblock Behavioral cloning from observation.
\newblock In \emph{IJCAI}, 2018.

\bibitem[Ng and Russell(2000)]{Ng_russell}
A.~Y. Ng and S.~J. Russell.
\newblock Algorithms for inverse reinforcement learning.
\newblock In \emph{ICML}, 2000.

\bibitem[Abbeel and Ng(2004)]{abbeel_ng}
P.~Abbeel and A.~Y. Ng.
\newblock Apprenticeship learning via inverse reinforcement learning.
\newblock In \emph{ICML}, 2004.

\bibitem[Zhou and Li(2018)]{wenchao_cav}
W.~Zhou and W.~Li.
\newblock Safety-aware apprenticeship learning.
\newblock In \emph{Computer Aided Verification {CAV}}, 2018.

\bibitem[{Cho} and {Oh}(2018)]{cho_mpc_stl}
K.~{Cho} and S.~{Oh}.
\newblock Learning-based model predictive control under signal temporal logic
  specifications.
\newblock In \emph{ICRA}, 2018.

\bibitem[{Memarian} et~al.(2020){Memarian}, {Xu}, {Wu}, {Wen}, and
  {Topcu}]{memarian_cdc2020}
F.~{Memarian}, Z.~{Xu}, B.~{Wu}, M.~{Wen}, and U.~{Topcu}.
\newblock Active task-inference-guided deep inverse reinforcement learning.
\newblock In \emph{CDC}, 2020.

\bibitem[Abel et~al.(2021)Abel, Dabney, Harutyunyan, Ho, Littman, Precup, and
  Singh]{abel2021expressivity}
D.~Abel, W.~Dabney, A.~Harutyunyan, M.~K. Ho, M.~L. Littman, D.~Precup, and
  S.~Singh.
\newblock On the expressivity of markov reward.
\newblock In \emph{NeurIPS}, 2021.

\bibitem[Everitt et~al.(2021)Everitt, Hutter, Kumar, and
  Krakovna]{everitt_reward_2021}
T.~Everitt, M.~Hutter, R.~Kumar, and V.~Krakovna.
\newblock Reward tampering problems and solutions in reinforcement learning: a
  causal influence diagram perspective.
\newblock \emph{Synthese}, May 2021.

\bibitem[Madumal et~al.(2020)Madumal, Miller, Sonenberg, and
  Vetere]{Madumal_Miller_Sonenberg_Vetere_2020}
P.~Madumal, T.~Miller, L.~Sonenberg, and F.~Vetere.
\newblock Explainable reinforcement learning through a causal lens.
\newblock \emph{AAAI}, 2020.

\bibitem[Juozapaitis et~al.(2019)Juozapaitis, Koul, Fern, Erwig, and
  Doshi-Velez]{xrl_reward_decompose}
Z.~Juozapaitis, A.~Koul, A.~Fern, M.~Erwig, and F.~Doshi-Velez.
\newblock Explainable reinforcement learning via reward decomposition.
\newblock In \emph{IJCAI. A Workshop on Explainable Artificial Intelligence.},
  2019.

\bibitem[El~Asri et~al.(2016)El~Asri, Piot, Geist, Laroche, and
  Pietquin]{score_irl}
L.~El~Asri, B.~Piot, M.~Geist, R.~Laroche, and O.~Pietquin.
\newblock Score-based inverse reinforcement learning.
\newblock In \emph{AAMAS}, 2016.

\bibitem[Bıyık et~al.(2022)Bıyık, Losey, Palan, Landolfi, Shevchuk, and
  Sadigh]{biyik_preference}
E.~Bıyık, D.~P. Losey, M.~Palan, N.~C. Landolfi, G.~Shevchuk, and D.~Sadigh.
\newblock Learning reward functions from diverse sources of human feedback:
  Optimally integrating demonstrations and preferences.
\newblock \emph{IJRR}, 2022.

\bibitem[Fainekos and Pappas(2009)]{fainekos_robustness_2009}
G.~E. Fainekos and G.~J. Pappas.
\newblock Robustness of temporal logic specifications for continuous-time
  signals.
\newblock \emph{Theoretical Computer Science}, 410\penalty0 (42), 2009.

\bibitem[Jak{\v s}i\'c et~al.(2018)Jak{\v s}i\'c, Bartocci, Grosu, Nguyen, and
  Ni{\v c}kovi\'c]{jaksic_quantitative_2018}
S.~Jak{\v s}i\'c, E.~Bartocci, R.~Grosu, T.~Nguyen, and D.~Ni{\v c}kovi\'c.
\newblock Quantitative monitoring of {{STL}} with edit distance.
\newblock \emph{Formal Methods in System Design}, 53\penalty0 (1), 2018.

\bibitem[Donz\'e and Maler(2010)]{donze_robust_2010}
A.~Donz\'e and O.~Maler.
\newblock Robust satisfaction of temporal logic over real-valued signals.
\newblock In \emph{International {{Conference}} on {{Formal Modeling}} and
  {{Analysis}} of {{Timed Systems}}}. {Springer}, 2010.

\bibitem[Dhonthi et~al.(2021)Dhonthi, Schillinger, Rozo, and Nardi]{smooth_stl}
A.~Dhonthi, P.~Schillinger, L.~D. Rozo, and D.~Nardi.
\newblock Study of signal temporal logic robustness metrics for robotic tasks
  optimization.
\newblock \emph{CoRR}, abs/2110.00339, 2021.

\bibitem[Donz{\'{e}}(2010)]{breach}
A.~Donz{\'{e}}.
\newblock Breach, {A} toolbox for verification and parameter synthesis of
  hybrid systems.
\newblock In \emph{Computer Aided Verification {CAV}}, 2010.

\bibitem[Ni{\v{c}}kovi{\'{c}} and Yamaguchi(2020)]{rtamt_stl}
D.~Ni{\v{c}}kovi{\'{c}} and T.~Yamaguchi.
\newblock Rtamt: Online robustness monitors from stl.
\newblock In \emph{Automated Technology for Verification and Analysis}, 2020.

\bibitem[Hasselt(2010)]{doubleqlrn}
H.~Hasselt.
\newblock Double q-learning.
\newblock In \emph{NIPS}, 2010.

\bibitem[Raffin et~al.(2021)Raffin, Hill, Gleave, Kanervisto, Ernestus, and
  Dormann]{stable_baselines3}
A.~Raffin, A.~Hill, A.~Gleave, A.~Kanervisto, M.~Ernestus, and N.~Dormann.
\newblock Stable-baselines3: Reliable reinforcement learning implementations.
\newblock \emph{Journal of Machine Learning Research}, 22\penalty0
  (268):\penalty0 1--8, 2021.
\newblock URL \url{http://jmlr.org/papers/v22/20-1364.html}.

\bibitem[Gao et~al.(2014)Gao, Vedula, Reiley, Ahmidi, Varadarajan, Lin, Tao,
  Zappella, B{\'e}jar, Yuh, Chen, Vidal, Khudanpur, and Hager]{jigsaws}
Y.~Gao, S.~Vedula, C.~Reiley, N.~Ahmidi, B.~Varadarajan, H.~C. Lin, L.~Tao,
  L.~Zappella, B.~B{\'e}jar, D.~Yuh, C.~C. Chen, R.~Vidal, S.~Khudanpur, and
  G.~Hager.
\newblock Jhu-isi gesture and skill assessment working set (jigsaws): A
  surgical activity dataset for human motion modeling.
\newblock In \emph{Modeling and Monitoring of Computer Assisted Interventions
  (M2CAI) – MICCAI Workshop}, 2014.

\end{thebibliography}

\clearpage
\section*{Appendix}
\appendix
\section{Signal Temporal Logic}
\label{app:stl}
\textit{Signal Temporal Logic (STL)} is a real-time logic, generally interpreted over a dense-time domain for signals whose values are from a continuous metric space (such as $\Reals^n$). The basic primitive in STL is a {\em signal predicate} $\mu$ that is a formula of the form $f(\vx(t)) > 0$, where $\vx(t)$ is the tuple $(state, action)$ of the demonstration $\vx$ at time $t$, and $f$ maps the signal domain $\domain = (S \times A)$ to $\Reals$. STL formulas are then defined recursively using Boolean combinations of sub-formulas, or by applying an interval-restricted temporal operator to a sub-formula.  The syntax of STL is formally defined as follows: $\varphi ::=  \mu \mid \neg \varphi \mid \varphi \wedge \varphi \mid \alw_{I} \varphi \mid \ev_{I} \varphi \mid \varphi \until_{I} \varphi$. Here, $I = [a,b]$ denotes an arbitrary time-interval, where $a,b\in\Reals^{\ge 0}$. The semantics of STL are defined over a discrete-time signal $\sig$ defined over some time-domain $\timedomain$. The Boolean satisfaction of a signal predicate is simply \textit{True} ($\top$) if the predicate is satisfied and \textit{False} ($\bot$) if it is not, the semantics for the propositional logic operators $\neg, \land$ (and thus $\lor, \rightarrow$) follow the obvious semantics. The following behaviors are represented by the temporal operators:
\begin{itemize}%[leftmargin=1em,nosep]
    \item At any time $t$, $\always_I(\varphi)$ says that $\varphi$ must hold for all samples in $t+I$.
    \item At any time $t$, $\eventually_I(\varphi)$ says that $\varphi$ must hold \textit{at
    least once} for samples in $t+I$.
    \item At any time $t$, $\varphi \until_I \psi$ says that $\psi$ must hold at some time $t'$ in $t+I$, and in $[t,t')$, $\varphi$ must hold at all times.
        % \item $\varphi \release_I \psi$ says that when $\varphi$ hold in $I$, $\psi$
    % should stop being true.
\end{itemize}

\begin{definition}[Quantitative Semantics for Signal Temporal Logic]%
\label{def:quantitative}
    Given an algebraic structure $(\oplus,\otimes,\top,\bot)$, we define the
    quantitative semantics for an arbitrary signal $\sig$ against an STL formula
    $\varphi$ at time $t$ as in \autoref{tab:stl_quant}.
	\begin{table}[htbp]
	\caption{Quantitative Semantics of STL}
	\label{tab:stl_quant}
	\centering
    \begin{tabular}{cc}
      \toprule
      $\varphi$ &  $\robustness{\varphi}{t}$ \\
      \midrule
      $\mathit{true}$/$\mathit{false}$ & $\top$/$\bot$ \\
      $\mu$                       & $f(\sig(t))$ \\
      $\neg \varphi$              & $-\robustness{\varphi}{t}$ \\
      
      $\varphi_1 \wedge \varphi_2$ &
      $\otimes(\robustness{\varphi_1}{t},\robustness{\varphi_2}{t})$ \\
      
      $\varphi_1 \vee \varphi_2$ & 
      $\oplus(\robustness{\varphi_1}{t},\robustness{\varphi_2}{t})$ \\
      
      $\alw_I(\varphi)$ &  $\otimes_{\tau\in t+I}(\robustness{\varphi}{\tau})$ \\
      
      $\ev_I(\varphi)$ &  $\oplus_{\tau\in t+I}(\robustness{\varphi}{\tau})$ \\

      $\varphi \until_I \psi$ & $\oplus_{\tau_1\in t+I}
      (\otimes(\robustness{\psi}{\tau_1},\otimes_{\tau_2\in[t,\tau_1)}(\robustness{\varphi}{\tau_2}))$ \\
      \bottomrule
      % \\
      
      % $\varphi \release_I \psi$ &  $\otimes_{\tau_1\in t+I}
      % (\oplus(\robustness{\psi}{\tau_1},\oplus_{\tau_2\in[t,\tau_1)}(\robustness{\varphi}{\tau_2}))$
      % \\
	\end{tabular}
\end{table}
\end{definition}

A signal satisfies an STL formula $\varphi$ if it is satisfied at time $t=0$. Intuitively, the quantitative semantics of STL represent the numerical distance of ``how far" a signal is away from the signal predicate. For a given requirement $\varphi$, a demonstration or policy $d$ that satisfies it is represented as $d \models \varphi$ and one that does not, is represented as $d \not\models \varphi$. In addition to the Boolean satisfaction semantics for STL, various researchers have proposed quantitative semantics for STL, \cite{fainekos_robustness_2009,jaksic_quantitative_2018} that compute the degree of satisfaction (or \textit{robust satisfaction values}) of STL properties by traces generated by a system. In this work, we use the following interpretations of the STL quantitative semantics: $\top = +\infty$, $\bot = -\infty$, and $\oplus = \max$, and $\otimes = \min$, as per the original definitions of robust satisfaction proposed in \cite{fainekos_robustness_2009,donze_robust_2010}.

Using these semantics allows a demonstration that satisfies a specification to have non-negative robustness (score) for that specification, and a demonstration that violates it will have a negative robustness (score). We use STL in our work because it offers a rich set of quantitative semantics (\autoref{def:quantitative}) that are suitable for formal analysis and reasoning of systems. The requirements defined with STL are grounded w.r.t. the actual description of the tasks/objectives. Furthermore, STL allows designers or users to specify constraints that evolve over time and define causal dependencies among tasks. Their semantics allow for the definition of non-Markovian rewards and accurately evaluating trajectories and policies for RL.

In our setting, a task can consist of multiple specifications. However, the robustness of each specification may lie on different scales. Consider for example, a driving scenario, where one specification concerns the speed of the vehicle, while another concerns the steering angle. Since the measurement scale of speed is significantly larger than angle (e.g., $60$ mph vs $1.6^\circ$), the robustness of the corresponding specifications also differs significantly. Furthermore, if the maximum robustness a car can achieve is 60 and 1.6 for the respective specifications, then directly performing summation on them would induce bias towards the speed specification. To avoid this bias, the robustness ranges need to be normalized. Some common normalization techniques are surveyed in \cite{smooth_stl}. We use the $tanh$ hyperbolic smoothing in our work to bound the robustness values.

\section{Derivations and Proofs}
\label{app:proofs}

\subsection{Space of all directed graphs}

In regard to the number of different orderings in extracting local graphs, given $n$ specifications, the number of permutations or arrangements is $n!$. For each permutation, there is 1 operator  from $op = \{>, =\}$ that can be placed in between any two specifications (e.g., $a > b$). The number of such ``places'' is $n-1$ and hence the number of operator arrangements for each permutation is $2^{n-1}$ or $|op|^{n-1}$ in general. However, we can observe that one of the arrangements for each permutation consists of the `=' operator appearing in all the ``places''. For example, $a = b = c$ is the same as the permutation $b = a = c$ and so on. Hence, all the $n!$ permutations share this common/redundant ordering, and so we need to remove all but one of them. Thus, the total number of unique orderings over all the permutations is $n! \cdot |op|^{n-1} - n! + 1 = n! \cdot [|op|^{n-1} - 1] + 1$. Following the use of directed graphs to reduce this search space, we first need to derive the number of possible directed graphs. For a directed graph without self-loops, there are 3 possible edge categories between any two nodes - no edge, incoming (outgoing from other) and outgoing (incoming to other). In the worst case, the maximum number of edges in a DAG is $n(n-1)/2$ edges and so the total number of possible directed graphs is $3^{n(n-1)/2}$. This includes all the cycles formed in the directed graph, so we then need to compute and subtract the number of cycles to obtain the actual space of DAGs. For a directed graph with $n$ vertices, a cycle comprises at least 3 vertices because we allow only 1 edge to exist between any two nodes. So, adding the number of cycles, we get:
\begin{align*}
\sum_{k=3}^{n} \binom{n}{k} &= \sum_{k=0}^{n} \binom{n}{k} - \sum_{k=0}^{2} \binom{n}{k} \\
&= 2^n - [\frac{n!}{0!(n-0)!} + \frac{n!}{1!(n-1)!} + \frac{n!}{2!(n-2)!}] \\
&= 2^n - [1 + n + n(n-1)/2]
\end{align*}
We can then reverse the edges and obtain another $2^n - [1 + n + n(n-1)/2]$ cycles, therefore the total number of cycles is twice this number $ = 2^{n+1} - (n^2 + n + 2)$. Finally, the number of valid directed graphs is  $3^{n(n-1)/2} - [2^{n+1} - (n^2 + n + 2)] = 3^{n(n-1)/2} - 2^{n+1} + n^2 + n + 2$, which is still exponential, but has eliminated the factorial component of the search space.

\subsection{Proof of Lemma and Theorem}
We provide proofs for the Lemma and Theorem stated in the main paper.

The lemma states: \textit{For a DAG, the weights associated with the nodes computed via \eqref{eqn:dag_weights}, are non-negative.}

\begin{equation*}
    w(\varphi) = |\Phi| - |ancestor(\varphi)|
\end{equation*}

\begin{proof}[Proof Sketch]
From the LfD-STL framework, the weights for specifications represented by the DAG nodes are given by \eqref{eqn:dag_weights}. We know that $|\Phi| = n$ and $ancestor(\varphi)$ is a set whose cardinality is non-negative. In a DAG, there are no cycles and hence $|ancestor(\varphi)|$ is an integer in $[0, n-1]$. By this equation, the minimum weight (i.e., worst-case) for any node representing a specification $\varphi$ occurs when that node is a leaf and all other $n-1$ nodes are its ancestors. Therefore, $w(\varphi) = |\Phi| - ancestor(\varphi) \implies w(\varphi) = n - (n-1) = 1 \geq 0$. Similarly, the maximum value of $w(\varphi)$ is $n$, i.e., there are no ancestors when $\varphi$ is one of the root nodes in the DAG. This non-negative nature of weights also holds true when the weights are normalized via a \texttt{softmax} function since it is used to represent a probability distribution that lies in the interval $[0, 1]$. 
\end{proof}

Using this lemma, we derive the proof for the theorem as described below.

The theorem states: \textit{For any two demonstrations $\demo_1$ and $\demo_2$ in an environment, the partial ordering $\demo_1 \preceq \demo_2$ is preserved by PeGLearn.}
\begin{proof}
Recall that for any two demonstrations $\demo_1$ and $\demo_2$ in an environment, if $\demo_1 \preceq \demo_2$, then the cumulative rating/scores are such that $r_{\demo_1} \leq r_{\demo_2}$. Also recall the notation that $\boldsymbol{\rho_{\demo_i}} = [\rho_{i1}, \cdots, \rho_{in}]^T$. Let there be $n$ specifications for the environment, then for these two demonstrations, we have:
\begin{align*}
\mathcal{Z} = \begin{bmatrix}
\rho_{11} & \rho_{12} & \cdots & \rho_{1n} \\
\rho_{21} & \rho_{22} & \cdots & \rho_{2n} \\
\end{bmatrix}
\end{align*}
and $\mathbf{w} = [w_1, w_2, ..., w_n]^T$. W.l.o.g., let $\demo_2$ be at least as good as $\demo_1$, i.e., we have $\rho_{1j} \leq \rho_{2j}, \forall j \in \{1, \cdots, n\}$. The cumulative scores for the demonstrations are $r_{\demo_i} =  \boldsymbol{\rho_{\demo_i}}^T \cdot \mathbf{w}$ where $i \in \{1, 2\}$.
%We know that $\rho_{1j} \leq \rho_{2j}, \forall j \in \{1, \cdots, n\}$. 
For any constant $w_j \geq 0$, 
\begin{align*}
& w_j \cdot \rho_{1j} \leq w_j \cdot \rho_{2j} \\
& \implies \sum_{j=1}^{n} w_j \cdot \rho_{1j} \leq \sum_{j=1}^{n} w_j \cdot \rho_{2j} \\
& \implies \boldsymbol{\rho_{\demo_1}}^T \cdot \mathbf{w} \leq \boldsymbol{\rho_{\demo_2}}^T \cdot \mathbf{w} \\
& \implies r_{\demo_1} \leq r_{\demo_2} \\
& \text{This holds iff } w_j \geq 0, \forall w_j \in \mathbf{w}
\end{align*}
 
Once the global DAG is learned, the weights for specifications (nodes) are computed via \eqref{eqn:dag_weights}. From the above Lemma, we have shown that these weights are all non-negative. Since the LfD-STL framework ranks the demonstrations by their cumulative scores, this guarantees that better demonstrations are always ranked higher than the others, i.e., a partial order is created, and also provide justification for the use of DAGs.
\end{proof}

\section{Additional Details on Experiments}
\label{app:extra_exp}

The STL formulas in our discrete-world and 2D driving experiments were specified and evaluated using Breach \cite{breach}, and the specifications for the MiR100 and CARLA experiments were evaluated in RTAMT-STL library \cite{rtamt_stl}. Note that the complexity of evaluating a trajectory w.r.t. a temporal logic specification is polynomial in the length of the signal and specification \cite{stl_complexity}. However, tools such as Breach and RTAMT are capable of producing linear-time complexity when evaluating In the 2D driving simulator experiment, we used the same neural network architecture as in our prior work \cite{puranic_ral2021} that was trained using $PyTorch$. All experiments were performed on a desktop machine with \textit{AMD Ryzen 7 3700X 8-core} CPU and \textit{Nvidia RTX 2070-Super} GPU.

\subsection{Discrete-World}

We use a grid environment, based on the OpenAI Gym $Frozenlake$ environment, consisting of a set of states $S = \{start, goals, obstacles\}$ of varying grid sizes such as: $5 \times 5$, $8 \times 8$ and $15 \times 15$ and randomizing the obstacle locations. Stochasticity in the range $p \in [0.1, 0.8]$ was introduced to the transition dynamics. This environment was created using $PyGame$ library where users provided demonstrations in the $PyGame$ GUI by clicking on their desired states with the task to reach the goal state from start without hitting any obstacles. Due to the stochasticity, {\em unaware to the users}, their clicked state may not always end up at the desired location. The user then proceeds to click from that unexpected state till they quit or reach the goal. Just as in \cite{puranic_corl2020, puranic_ral2021}, we used \textit{Manhattan} distance as the distance metric and formulated the STL specifications:

\begin{enumerate}
	\item Avoid obstacles at all times (hard requirement): $\mathbf{ \varphi_1 := G_{[0,T]} (d_{obs}[t] \ge 1) }$, where $T$ is the length of a demonstration and $d_{obs}$ is the minimum distance of robot from obstacles computed at each step $t$.
	\item Eventually, the robot reaches the goal state (soft requirement): $\mathbf{ \varphi_2 := F_{[0,T]} (d_{goal}[t] < 1) }$, where $d_{goal}$ is the distance of robot from goal computed at each step. $\varphi_2$ depends on $\varphi_1$.
	\item Reach the goal as fast as possible (soft requirement): $\mathbf{ \varphi_3 := F_{[0,T]} (t \le T_{goal}) }$, where $T_{goal}$ is the upper bound of time required to each the goal, which is computed by running breadth-first search algorithm from start to goal state, since the shortest policy must take at least $T_{goal}$ to reach the goal. $\varphi_3$ depends on both $\varphi_1$ and $\varphi_2$ in the DAG.
\end{enumerate}

The PeGLearn algorithm was evaluated against (i) user-defined DAGs, (ii) MaxEntropy IRL, and (iii) MaxCausalEntropy IRL. Once rewards were extracted from each algorithm for all environment settings, we used Double Q-Learning \cite{doubleqlrn}, as it is suited for stochastic settings, with the modifications to the algorithm at 2 steps (reward update and termination) as described by \cite{puranic_corl2020}. The number of episodes varied based on environment complexity such as grid size, number and locations of obstacles. The discount factor $\gamma$ was set to 0.8 and $\epsilon$-greedy strategy with decaying $\epsilon$ for actions was used. A learning rate of $\alpha = 0.1$ was found to work reasonably well after analyzing hyperparameters. Our evaluations over $100$ trials showed that policies independently learned from PeGLearn and manually-defined DAGs were able to achieve a task success rate of $80\%$ and $81\%$ respectively for the environments with $0.2$ stochasticity. However, the execution time of PeGLearn was within a 2-second increment over that of LfD-STL with manually-specified DAGs.

\subsection{2D Driving Simulator}

For this experiment, we used the same kinematic model for a car, described by the following equations:

\begin{align*}
\dot{x} &= v \cdot cos(\theta) + \mathcal{N}(0, \sigma^2); \
\dot{y} = v \cdot sin(\theta) + \mathcal{N}(0, \sigma^2) \\
\dot{v} &= u_1 \cdot u_2; \
\dot{\theta} = v \cdot tan(\psi); \
\dot{\psi} = u_3
\end{align*}

where $x$ and $y$ represent the $XY$ position of the car; $\theta$ is the heading; $v$ is the velocity; $u_1$ is the input acceleration; $u_2$ is the gear indicating forward (+1) or backward (-1); $u_3$ is the input to steering angle $\psi$. The state of the car at time $t$ is given by $S_t=[x, y, \theta, v, \dot{x}, \dot{y}, \dot{\theta}, \dot{v}]^T$. Demonstrations are provided by users via an analog \textit{Logitech G29} steering with pedal controller or via keyboard inputs. For comparison with prior work, we utilized the same 8 (6 good and 2 bad) demonstrations recorded earlier (\autoref{fig:dubins_demos}).

\begin{figure}[tb]
\centering
\includegraphics[scale=0.3]{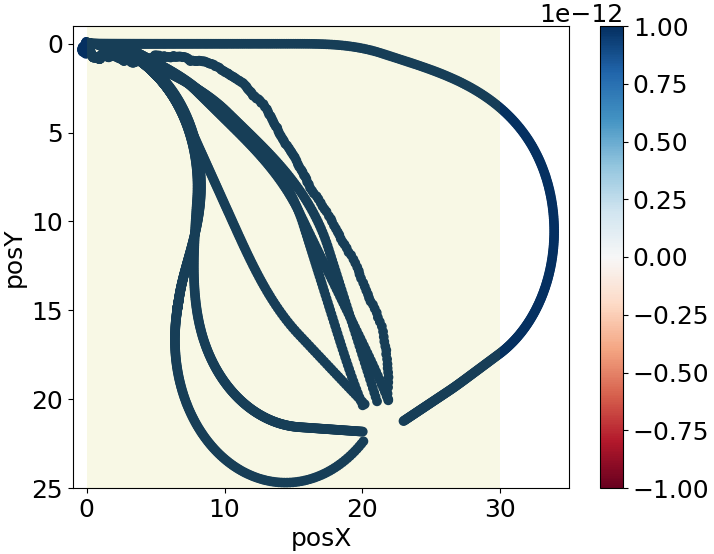}
\caption{Demonstrations collected for the 2D car simulator.}
\label{fig:dubins_demos}
\end{figure}

The distance metric used in this space is $Euclidean$. The specifications for this scenario are as follows:

\begin{enumerate}
	\item Avoid obstacles at all times (hard requirement): $\mathbf{ \varphi_1 := G_{[0,T]} (d_{obs}[t] \ge dSafe) }$, where $T$ is the length of a demonstration and $d_{obs}$ is the minimum distance of the car from $\mathcal{H}$ computed at each step $t$. For our experiments, we used $\mathtt{dSafe} = 3$ units.
	\item Always stay within the workspace/drivable region (hard requirement): $\mathbf{ \varphi_2 := G_{[0,T]} ((x,y) \in Box(30, 25))}$, where the workspace is defined by a rectangle of dimensions $30 \times 25$ square units. The $Box$ is an indicator for the real-valued data in the $OpenAI Gym$ library.
	\item Eventually, the robot reaches the goal state (soft requirement): $\mathbf{ \varphi_3 := F_{[0,T]} (d_{goal}[t] < \delta) }$, where $d_{goal}$ is the distance between centers of car and goal computed at each step $t$ and $\delta$ is a small tolerance when the center of the car is ``close enough'' to the goal's center. $\varphi_3$ depends on $\varphi_1$ and $\varphi_2$ in the DAG.
\end{enumerate}

The rewards are assigned to states by modeling the states $s$ as samples of multi-variate Gaussian distribution $\mathcal{N}(\mu, \sigma^2 I)$ where $\mu = s$ and $\sigma$ represents the deviations in noise levels, that can be tuned. Here, we use $\sigma=0.03$. For each $s$, we generated $k=20$ samples to represent the reachable set and assigned stochastic rewards as described in \cite{puranic_ral2021}. The neural network used for regressing the rewards consisted of 2 layers with 200 neurons in each layer that were activated by ReLU. The reward inference for both PeGLearn and manual-DAG baseline had execution times of less than {\em 30 seconds}.

\subsection{MiR100 Navigation Experiment}
Given the 30 demonstrations, PeGLearn was able to extract the rewards within {\em 30 seconds}. The expert rewards that were used for this task comprised of the following components: (i) $Euclidean$ distance between the robot and goal, (ii) power used by the robot motors for linear and angular velocities, (iii) distance to obstacles to detect collisions, and (iv) an optional penalty if the robot exited the environment boundary walls. The PPO agent that was trained separately on the expert and PeGLearn rewards used the default architecture and hyperparameter settings from \cite{stable_baselines3}. Both training sessions were run for 3e6 steps, with each session lasting about 30 hours on our hardware. Likewise, the D4PG agent was trained independently on each reward function, under similar training conditions (hyperparameters shown in \autoref{tab:d4pg_params}), with each training session lasting about 12 hours. Each of these RL agents were then evaluated on 100 test runs (trials) to compute the success rates.

\begin{table}[!ht]
	\caption{D4PG hyperparameters.\label{tab:d4pg_params}}
	\centering
    \begin{tabular}{cc}
      \toprule
      Hyperparameter &  Value \\
      \midrule
      Actors & 5 \\
      Learners & 1 \\
      Actor MLP & 256 $\rightarrow$ 256 \\
      Critic MLP & 256 $\rightarrow$ 256 \\
      N-Step & 5 \\
      Atoms & 51 \\
      $V_{min}$ & -10 \\
      $V_{max}$ & 10 \\
      Exploration Noise & 0.3 \\
      Discount Factor & 0.99 \\
      Mini-batch Size & 256 \\
      Actor Learning Rate & 5e-4 \\
      Critic Learning Rate & 5e-4 \\
      Memory Size & 1e6 \\
      Learning Batch & 64 \\
      \bottomrule
	\end{tabular}
\end{table}

\subsection{CARLA-AMT Survey}

The demonstrations for this experiment used the same analog hardware for the 2D car simulator (\autoref{fig:carla_teleop}).
\begin{figure}
\centering
\includegraphics[scale=0.2]{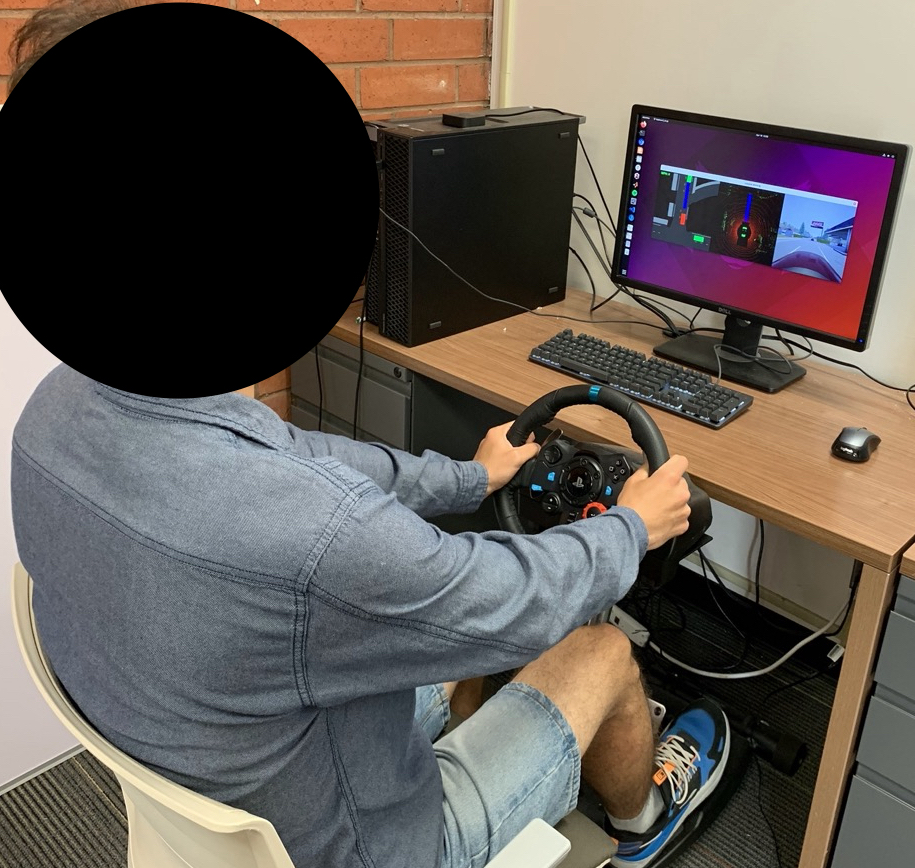}
\caption{Teleoperation demonstrations in CARLA.}
\label{fig:carla_teleop}
\end{figure}
The formal description of the STL task specifications for the CARLA simulator are as follows:
\begin{enumerate}

	\item Keeping close to the center of the lane:\\ $\mathbf{ \varphi_1 := G_{[0,T]} (d_t^{lane} \le \delta) }$, where $T$ is the length of a demonstration, $d_t^{lane}$ is the distance of car from the center of the lane at each step $t$ and $\delta$ is a small tolerance factor. The width of a typical highway lane in the US is 12 ft (3.66 m) \footnote{Based on USDOT highway \label{ft_us} and US vehicle specifications (e.g., Ford F-150).} and the average width of a big vehicle (e.g., SUV or pickup truck) is 7 ft (2.13 m) \footref{ft_us}, which leaves about 2.5 ft (0.76 m) of room on either side of the vehicle. Hence, we chose to use 1 ft (0.3 m) as the tolerance factor to accurately track the lane center while also providing a small room for error.
	
	\item Maintaining speed limits:\\ $\mathbf{ \varphi_2 := G_{[0,T]} (v_{min} \leq v_t \leq v_{max}) }$, where $v_t$ is speed of the ego/host car at each timestep $t$, and $v_{min}$ and $v_{max}$ are the speed limits. Since it is a US highway scenario, the $v_{max} = 65$ mph and $v_{min} =  0$ mph.
%	$\mathbf{ \varphi_2 := G_{[0,T]} ((v[t] \ge v_{min}) \wedge (v[t] \le v_{max})) }$
	
	\item Maintaining safe distance from any lead vehicle:\\ $\mathbf{ \varphi_3 := G_{[0,T]} (safety\_flag_t \le 0) }$, \\ where $\mathtt{safety\_flag_t}$ is a binary signal that outputs 0 if the ego is safe (i.e., there is no vehicle directly in front of the ego in the same lane whose distance is closer than some threshold $d_{safe}$) and 1 otherwise. In OpenAI Gym-CARLA, the safe distance was set to \texttt{15 m}.
	
\end{enumerate}

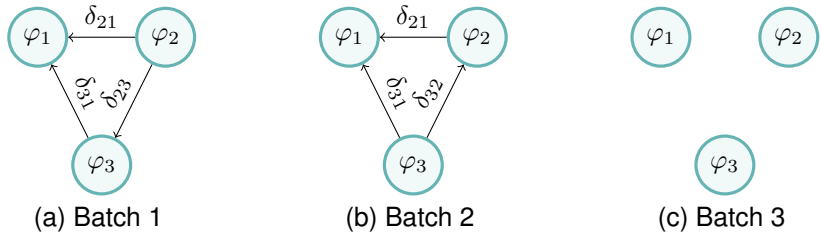
\begin{figure}[htbp]
\centering
\subfloat[Batch 1]{
\begin{tikzpicture}[scale=1.0, transform shape, node distance=1.7cm]
	%Nodes
%			
	\node[roundnode] at (0,0) (phi1) {$\varphi_1$};
	\node[roundnode] (phi2) [right of= phi1] {$\varphi_2$};
	\coordinate (Middle) at ($(phi1)!0.5!(phi2)$);
%		\node[roundnode] (phi3) [below of=phi1,xshift=.75cm] {$\varphi_3$};
	\node[roundnode] (phi3) [below of=Middle] {$\varphi_3$};
	
	%Lines
	\draw[->] (phi2) -- node[midway, above, sloped] {$\delta_{21}$} (phi1);
	\draw[->] (phi3) -- node[midway, above, sloped] {$\delta_{31}$} (phi1);
	\draw[->] (phi2) -- node[midway, above, sloped] {$\delta_{23}$} (phi3);
\end{tikzpicture}} \hfil
\subfloat[Batch 2]{
\begin{tikzpicture}[scale=1.0, transform shape, node distance=1.7cm]
	%Nodes
%			
	\node[roundnode] at (0,0) (phi1) {$\varphi_1$};
	\node[roundnode] (phi2) [right of= phi1] {$\varphi_2$};
	\coordinate (Middle) at ($(phi1)!0.5!(phi2)$);
%		\node[roundnode] (phi3) [below of=phi1,xshift=.75cm] {$\varphi_3$};
	\node[roundnode] (phi3) [below of=Middle] {$\varphi_3$};
	
	%Lines
	\draw[->] (phi2) -- node[midway, above, sloped] {$\delta_{21}$} (phi1);
	\draw[->] (phi3) -- node[midway, above, sloped] {$\delta_{31}$} (phi1);
	\draw[->] (phi3) -- node[midway, above, sloped] {$\delta_{32}$} (phi2);
\end{tikzpicture}} \hfil
\subfloat[Batch 3]{
\begin{tikzpicture}[scale=1.0, transform shape, node distance=1.7cm]
	%Nodes
%			
	\node[roundnode] at (0,0) (phi1) {$\varphi_1$};
	\node[roundnode] (phi2) [right of= phi1] {$\varphi_2$};
	\coordinate (Middle) at ($(phi1)!0.5!(phi2)$);
	\node[roundnode] (phi3) [below of=Middle] {$\varphi_3$};
	
	%Lines
\end{tikzpicture}} \hfil
%\subfloat[Teleoperation]{\includegraphics[width=1.4in, height=1.2in]{carla/carla_demo1_blur}}
\caption{DAGs for the CARLA simulator experiment.}
\label{fig:carla_graphs}
\end{figure}

For the online AMT survey, we initially recruited 150 human participants and took numerous measures to ensure reliability of results. We posed a control question at the end to test their attention to the task, and eliminated data associated with the wrong answer, including incomplete data, resulting in 146 samples. All participants had an approval rating over $98\%$ and the demographics are as follows: (i) 73 males, 72 females, 1 other, (ii) participant age ranged from 22 to 79 with an average age of 40.67, and (iii) average driving experience of 22.4 years. Our survey collected the following information from each participant:
\begin{itemize}[leftmargin=1em,nosep]
\item Participant information: Number of years of driving experience, age, gender and experience with video games.
\item Ratings on a scale of 1 (worst) - 5 (best) for the queries/specifications: (i) driver staying close to the lane center, (ii) driver maintaining safe distance to lead vehicle(s) and (iii) driver respecting speed limits of the highway.
\item Ratings on a scale of 1 (lowest) - 3 (highest) on the overall driving behavior shown in these 5 videos and also how the participants would prioritize each of the specifications if they were driving in that scenario.
\end{itemize}

The web-based questionnaire showed a batch of 5 videos to a participant, where each video was accompanied by the three questions regarding the task specifications. A still of one of the videos is shown in \autoref{fig:survey_vid}. Users were presented with a dropdown menu in which each option was a $Likert$ rating from 1 (lowest) to 5 (highest). We then presented users with a question to rate the overall behavior of all 5 videos in the batch w.r.t. the task specifications on a scale of 1 (lowest) to 3 (highest). Finally, a control question was posed regarding the color of the car shown in the videos to test the attention of the users, since the colors of cars were same across all 15 videos. The graphs for each batch obtained via PeGLearn is shown in \autoref{fig:carla_graphs}.

\begin{figure}[tb]
\includegraphics[width=\linewidth]{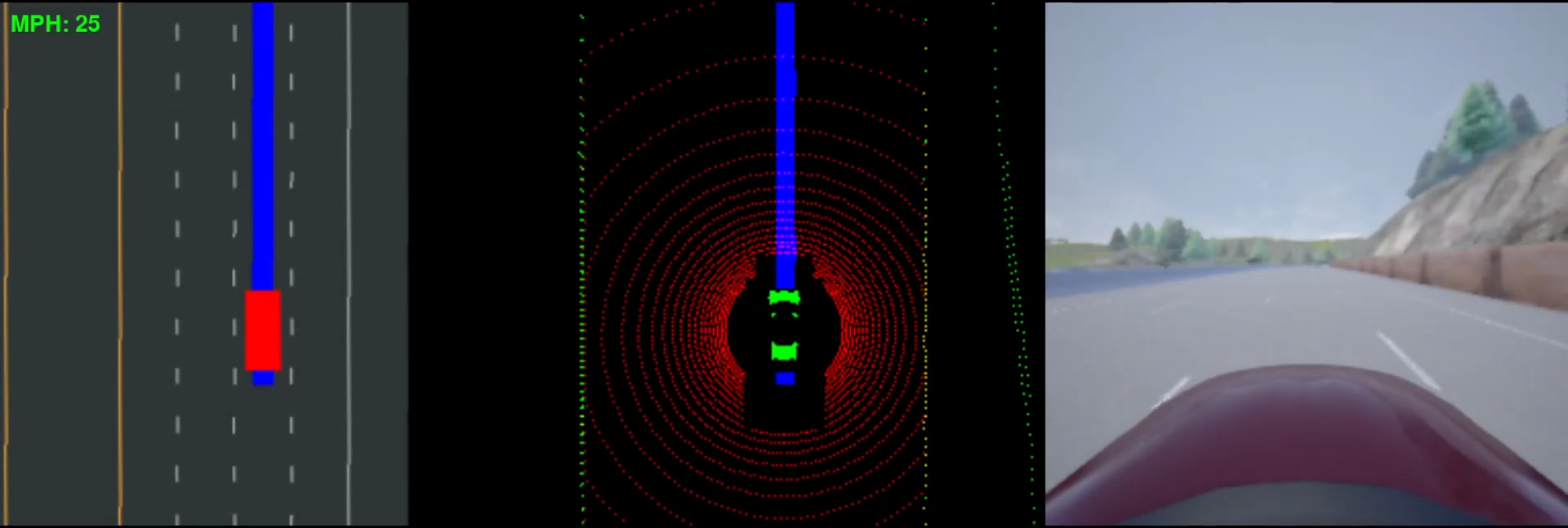}
\caption{A frame from one of the survey videos.}
\label{fig:survey_vid}
\end{figure}

The overall orderings from human experts and our PeGLearn algorithm for each AMT survey batch are shown in \autoref{fig:amt_batch_summaries}. To compare the ratings, we first normalized all the human and PeGLearn ratings to be in the range $[0, 3]$. The $user$ bars correspond to the human expert ratings while $auto$ represents our algorithm rating, which is deterministic and hence there are no error bars.

\begin{figure*}[tb]
\centering
\subfloat[Batch A\label{fig:carla_batchA}]{\includegraphics[height=4cm, width=4.5cm]{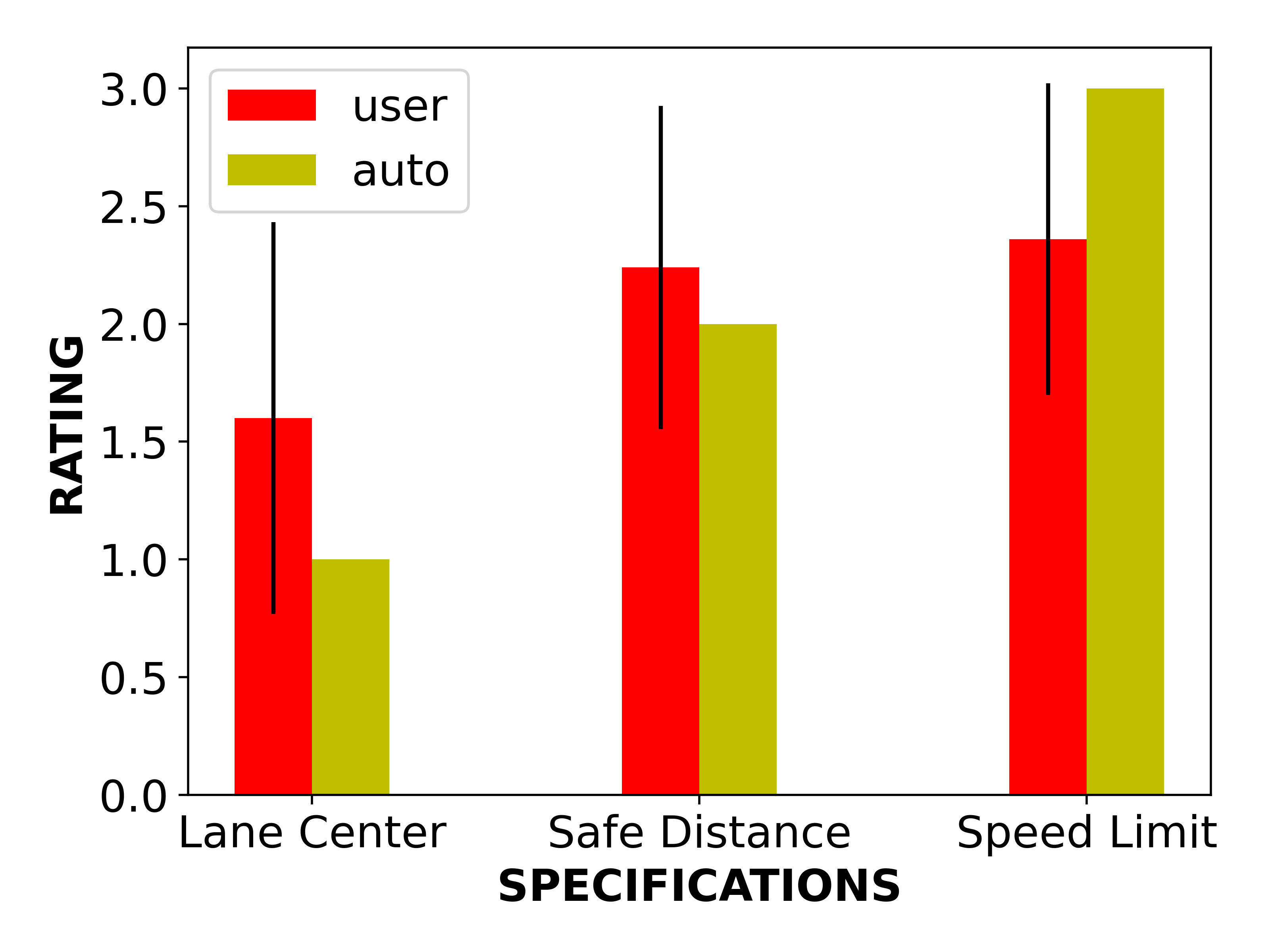}} \hfil
\subfloat[Batch B\label{fig:carla_batchB}]{\includegraphics[height=4cm, width=4.5cm]{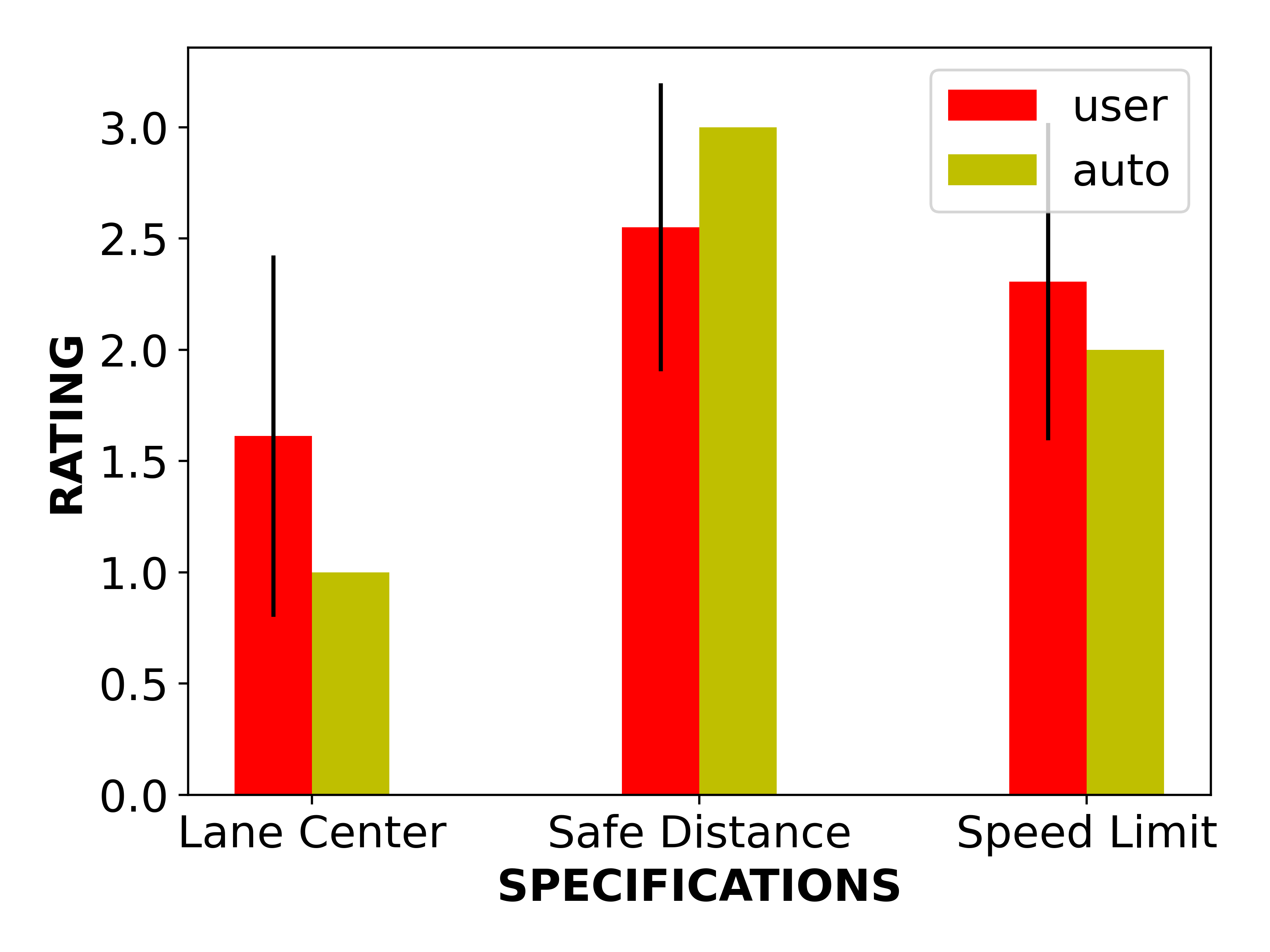}} \hfil
\subfloat[Batch C\label{fig:carla_batchC}]{\includegraphics[height=4cm, width=4.5cm]{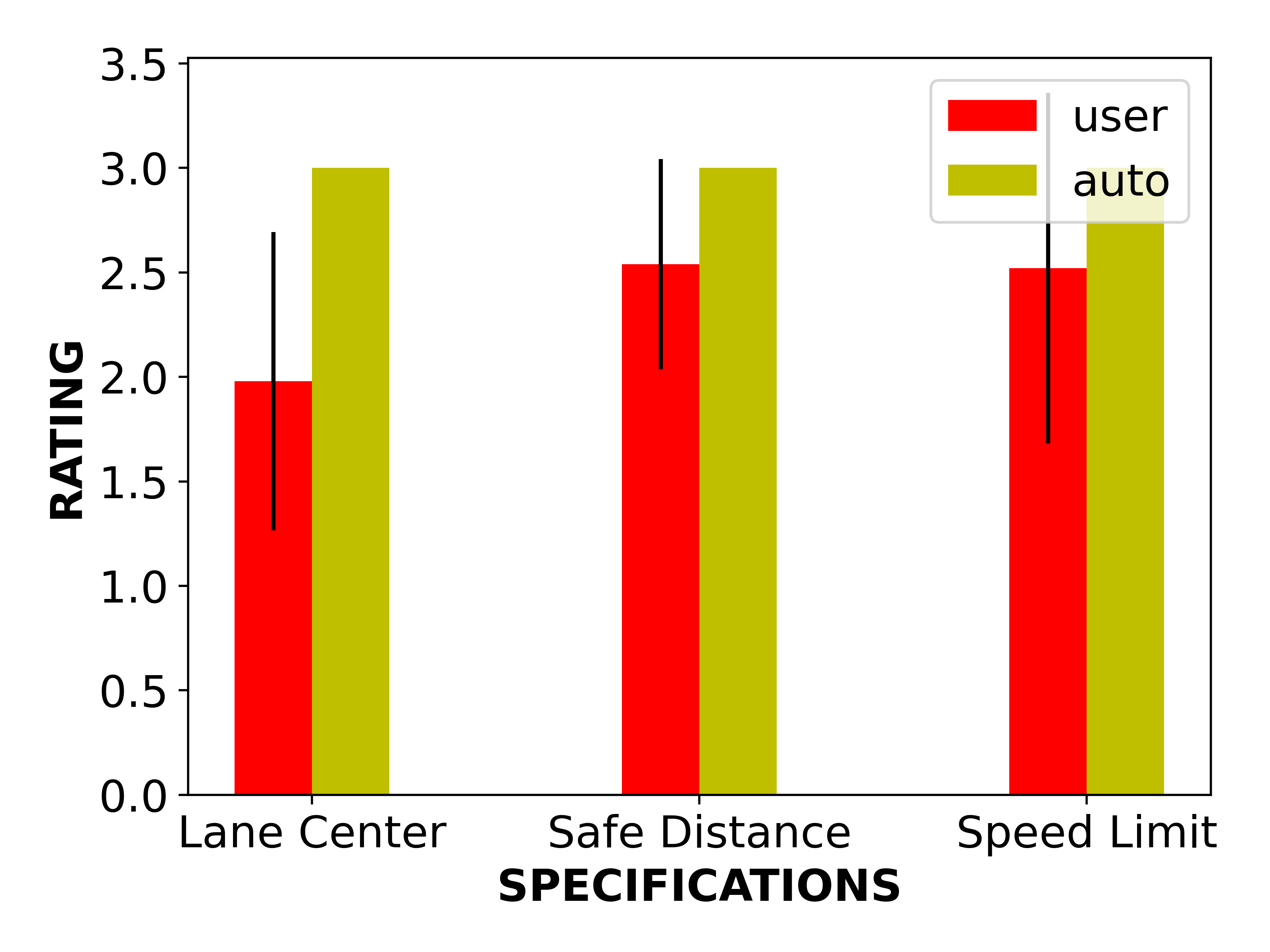}}
\caption{Comparison of specification orderings between humans and PeGLearn.}
\label{fig:amt_batch_summaries}
\end{figure*}

\section{JIGSAWS (Surgical Robot Dataset)}
\label{app:jigsaws}
To show the generalizability of our method to performance assessment metrics beyond those induced by temporal logics, we evaluated it on human ratings (e.g., $Likert$ scale) provided for various robotic surgical tasks \cite{jigsaws} performed on a {\em Da Vinci Surgical robot system}. As described in the dataset, an expert surgeon provided evaluations or ratings for 8 different surgeons with various expertise levels on 3 basic surgical tasks - knot-tying, needle-passing and suturing. There are 6 specifications for evaluating the performance of surgeons on the tasks and the ratings are measured on a scale from 1 (lowest) to 5 (highest) for each specification. The 6 specifications or evaluation criteria are:
\begin{enumerate}
\item Respect for tissue (TR) - force exerted on tissue.
\item Suture/needle handling (SNH) - control while tying knots.
\item Time and motion (TM) - fluent motions and efficient.
\item Flow of operation (FO) - planned approach with minimum interruptions between moves.
\item Overall performance (P).
\item Quality of final product (Q).
\end{enumerate}

Using this rating scheme, our method was able to generate a performance-graph for each class of expertise as shown in \autoref{fig:jigsaws} for the \textit{knot-tying} task. If the expertise levels were unknown, then the generated graph would be as indicated in \autoref{fig:jigsaws_agg}. We obtained similar performance graphs for the other 2 surgical tasks - \textit{needle-passing} and \textit{suturing}. This shows how human evaluations can be used in environments such as these where it is difficult even for experts to express the tasks in a formal language. From the graphs, we can see that all 3 categories of surgeons showed maximum performance on (Q) and (TM). However, there were differences in the other specifications. For example, experts had a higher rating of (P) over (TR) compared to intermediates. One possible explanation for this is that experts typically perform multiple consecutive surgeries, and so they optimize on the (P) and (FO) aspects compared to (TR), while the intermediate-level surgeons are trainees who are still learning the nuances of surgeries and are focusing more on the qualitative aspects such as (TR) over quantity and speed. Similar reasoning can be applied to each category of surgeons using these DAGs.

\begin{remark}
We acknowledge that providing individual ratings for every demonstration-specification pair is tedious since the complexity of manually specifying the performance graph is exponential as elaborated in \appref{app:proofs}. This is because one needs to take into account, not just the labels or ratings, but also the orderings (permutations) among those labels. In other words, a user needs to assign $\mathcal{O}(mn)$ ratings and also compare them with different permutations of those ratings, i.e., creating relative priorities to specify the graph, which is exponential in $O(n^2)$. Thus, manually defining the graphs results in a very large complexity as shown. Our method significantly eliminates this complex manual labor by using only minimal inputs from users as it is much easier to provide individual labels than having to compare with all permutations of the labels. To even further reduce human inputs, a potential solution we will consider for future work is to use deep temporal learning methods to learn from existing labeled data and predict the labels for newer demonstrations. We argue that some form of human feedback would be necessary to provide formal guarantees in learning rewards since it provides a ground truth baseline.
\end{remark}

\begin{figure*}[tb]
\centering
\subfloat[Experts]{
	\begin{tikzpicture}[scale=0.8, transform shape, node distance=2cm,
	roundnode/.style={circle, draw=teal!60, fill=teal!5, very thick, minimum size=10mm}
	]
		%Nodes
%		\node[roundnode](phi2){$\varphi_2$};
%		\node[roundnode](phi3)[right=of phi2]{$\varphi_3$};
%		\node[roundnode](phi1)[below right=of phi2]{$\varphi_1$};
		
%		\node[roundnode](phi1){$\varphi_1$};
%		\node[roundnode](phi3)[above right=of phi1]{$\varphi_3$};
%		\node[roundnode](phi2)[above left=of phi1]{$\varphi_2$};
		
		\node[roundnode] (1) {TR};
		\node[roundnode] (2)[right=of 1] {SNH};
		\node[roundnode] (3)[below=of 1] {TM};
		\node[roundnode] (4)[below=of 2] {FO};
		\coordinate (MiddleL) at ($(1)!0.5!(3)$);
		\node[roundnode] (6)[left of=MiddleL] {Q};
		\coordinate (MiddleR) at ($(2)!0.5!(4)$);
		\node[roundnode] (5)[right of=MiddleR] {P};
		
		%Lines
		\draw[->] (1) -- (3);
		
		\draw[->] (2) -- (3);
		\draw[->] (2) -- (1);
		
		\draw[->] (4) -- (3);
		\draw[->] (4) -- (2);
		\draw[->] (4) -- (1);
		\draw[->] (4) -- (5);
		\draw[->] (4) -- (3);
		
		\draw[->] (5) -- (3);
		\draw[->] (5) -- (2);
		\draw[->] (5) -- (1);
		
		\draw[->] (6) -- (1);
		\draw[->] (6) -- (2);
		\draw[->] (6) -- (3);
		\draw[->] (6) -- (4);
		\draw[->] (6) -- (5);
	\end{tikzpicture}
	} \hfil
\subfloat[Intermediates]{	
	\begin{tikzpicture}[scale=0.8, transform shape, node distance=2cm,
	roundnode/.style={circle, draw=teal!60, fill=teal!5, very thick, minimum size=10mm}
	]
		%Nodes
%		\node[roundnode](phi2){$\varphi_2$};
%		\node[roundnode](phi3)[right=of phi2]{$\varphi_3$};
%		\node[roundnode](phi1)[below right=of phi2]{$\varphi_1$};
		
%		\node[roundnode](phi1){$\varphi_1$};
%		\node[roundnode](phi3)[above right=of phi1]{$\varphi_3$};
%		\node[roundnode](phi2)[above left=of phi1]{$\varphi_2$};
		
		\node[roundnode] (1) {TR};
		\node[roundnode] (2)[right=of 1] {SNH};
		\node[roundnode] (3)[below=of 1] {TM};
		\node[roundnode] (4)[below=of 2] {FO};
		\coordinate (MiddleL) at ($(1)!0.5!(3)$);
		\node[roundnode] (6)[left of=MiddleL] {Q};
		\coordinate (MiddleR) at ($(2)!0.5!(4)$);
		\node[roundnode] (5)[right of=MiddleR] {P};
		
		%Lines
		\draw[->] (1) -- (5);
		\draw[->] (1) -- (4);
		\draw[->] (1) -- (3);
		\draw[->] (1) -- (2);
		
		\draw[->] (2) -- (3);
		
		\draw[->] (4) -- (5);
		\draw[->] (4) -- (3);
		\draw[->] (4) -- (2);
		
		\draw[->] (5) -- (3);
		\draw[->] (5) -- (2);
		
		\draw[->] (6) -- (1);
		\draw[->] (6) -- (2);
		\draw[->] (6) -- (3);
		\draw[->] (6) -- (4);
		\draw[->] (6) -- (5);
	\end{tikzpicture}
	}

\subfloat[Novices]{
	\begin{tikzpicture}[scale=0.8, transform shape, node distance=2cm,
	roundnode/.style={circle, draw=teal!60, fill=teal!5, very thick, minimum size=10mm}
	]
		%Nodes
%		\node[roundnode](phi2){$\varphi_2$};
%		\node[roundnode](phi3)[right=of phi2]{$\varphi_3$};
%		\node[roundnode](phi1)[below right=of phi2]{$\varphi_1$};
		
%		\node[roundnode](phi1){$\varphi_1$};
%		\node[roundnode](phi3)[above right=of phi1]{$\varphi_3$};
%		\node[roundnode](phi2)[above left=of phi1]{$\varphi_2$};
		
		\node[roundnode] (1) {TR};
		\node[roundnode] (2)[right=of 1] {SNH};
		\node[roundnode] (3)[below=of 1] {TM};
		\node[roundnode] (4)[below=of 2] {FO};
		\coordinate (MiddleL) at ($(1)!0.5!(3)$);
		\node[roundnode] (6)[left of=MiddleL] {Q};
		\coordinate (MiddleR) at ($(2)!0.5!(4)$);
		\node[roundnode] (5)[right of=MiddleR] {P};
		
		%Lines
		\draw[->] (1) -- (3);
		
		\draw[->] (2) -- (3);
		\draw[->] (2) -- (5);
		\draw[->] (2) -- (1);
		
		\draw[->] (4) -- (3);
		\draw[->] (4) -- (2);
		\draw[->] (4) -- (1);
		\draw[->] (4) -- (5);
		
		\draw[->] (5) -- (3);
		
		\draw[->] (6) -- (1);
		\draw[->] (6) -- (2);
		\draw[->] (6) -- (3);
		\draw[->] (6) -- (4);
		\draw[->] (6) -- (5);
	\end{tikzpicture}
	} \hfil
\subfloat[Cumulative \label{fig:jigsaws_agg}]{
	\begin{tikzpicture}[scale=0.8, transform shape, node distance=2cm,
	roundnode/.style={circle, draw=teal!60, fill=teal!5, very thick, minimum size=10mm}
	]
		%Nodes
%		\node[roundnode](phi2){$\varphi_2$};
%		\node[roundnode](phi3)[right=of phi2]{$\varphi_3$};
%		\node[roundnode](phi1)[below right=of phi2]{$\varphi_1$};
		
%		\node[roundnode](phi1){$\varphi_1$};
%		\node[roundnode](phi3)[above right=of phi1]{$\varphi_3$};
%		\node[roundnode](phi2)[above left=of phi1]{$\varphi_2$};
		
		\node[roundnode] (1) {TR};
		\node[roundnode] (2)[right=of 1] {SNH};
		\node[roundnode] (3)[below=of 1] {TM};
		\node[roundnode] (4)[below=of 2] {FO};
		\coordinate (MiddleL) at ($(1)!0.5!(3)$);
		\node[roundnode] (6)[left of=MiddleL] {Q};
		\coordinate (MiddleR) at ($(2)!0.5!(4)$);
		\node[roundnode] (5)[right of=MiddleR] {P};
		
		%Lines
		\draw[->] (1) -- (3);
		\draw[->] (1) -- (2);
		
		\draw[->] (2) -- (3);
		
		\draw[->] (4) -- (3);
		\draw[->] (4) -- (2);
		\draw[->] (4) -- (1);
		\draw[->] (4) -- (5);
		
		\draw[->] (5) -- (3);
		\draw[->] (5) -- (2);
		
		\draw[->] (6) -- (1);
		\draw[->] (6) -- (2);
		\draw[->] (6) -- (3);
		\draw[->] (6) -- (4);
		\draw[->] (6) -- (5);
	\end{tikzpicture}
	}
\caption{DAGs for the Knot-Tying task. (a)-(c) DAG for each level of expertise: Experts, Intermediates and Novices respectively. (d) DAG for all surgeons, without discriminating expertise levels.}
\label{fig:jigsaws}
\end{figure*}
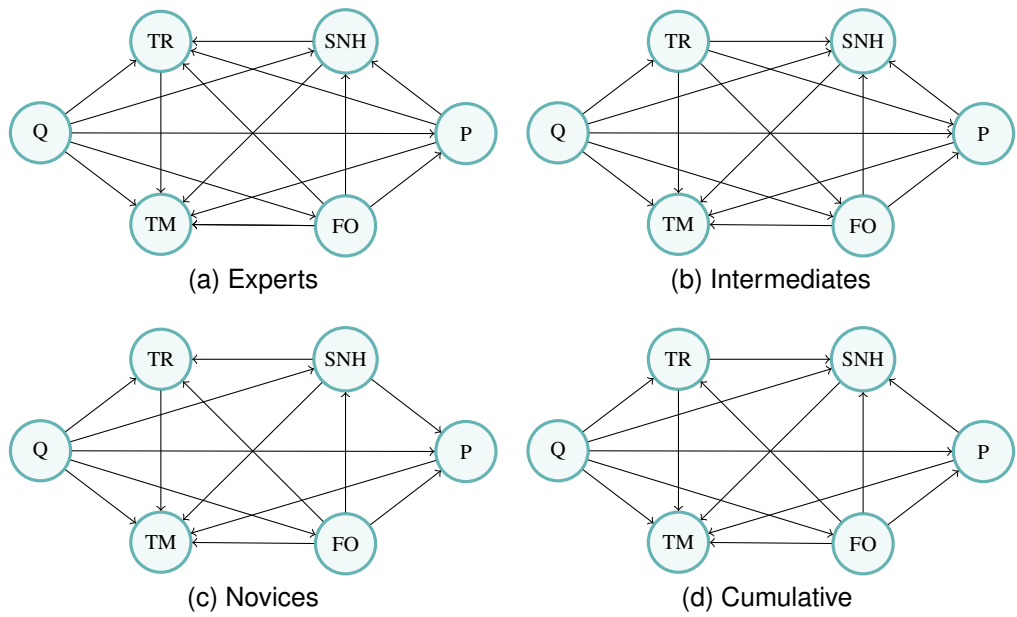

\end{document}